\documentclass{article}

\usepackage{arxiv}

\usepackage[utf8]{inputenc} 
\usepackage[T1]{fontenc}    
\usepackage{hyperref}       
\usepackage{url}            
\usepackage{booktabs}       
\usepackage{amsfonts}       
\usepackage{amsmath}
\usepackage{nicefrac}       
\usepackage{microtype}      
\usepackage{lipsum}
\usepackage{graphicx}
\usepackage{cancel}

\usepackage{natbib} 

\usepackage{amssymb}
\usepackage{mathrsfs}

\usepackage{xcolor}

\usepackage[titlenumbered,ruled,linesnumbered]{algorithm2e}
\usepackage[noend]{algpseudocode}

\title{{A general linear-time inference method for Gaussian Processes on one dimension}}

\author{
Jackson Loper \\
 Data Science Institute \\
 Columbia University \\
 New York, New York 10027\\
 \texttt{jl5116@columbia.edu}\\
 \And
 David Blei \\
 Data Science Institute \\
 Departments of Statistics and Computer Science\\
 Columbia University \\
 New York, New York 10027\\
 \And
 John P. Cunningham \\
 Department of Neuroscience\\
 Mortimer B. Zuckerman Mind Brain Behavior Institute \\
 Grossman Center for the Statistics of Mind \\
 Columbia University \\ 
 \And
 Liam Paninski \\
 Departments of Statistics and Neuroscience\\
 Mortimer B. Zuckerman Mind Brain Behavior Institute \\
 Grossman Center for the Statistics of Mind \\
 Columbia University \\ 
}

\usepackage{amsthm}
\newtheorem{lemma}{Lemma}
\newtheorem{theorem}{Theorem}
\theoremstyle{definition}
\newtheorem{definition}{Definition}

\newtheorem{prop}{Proposition}

\newcommand{\diffusion}{N}
\newcommand{\rotparm}{R}
\newcommand{\latV}{\mathbf{z}}
\newcommand{\obsV}{\mathbf{x}}
\newcommand{\brownianmotion}{\mathbf{w}}
\newcommand{\LEG}{\mathrm{LEG}}
\newcommand{\reals}{\mathbb{R}}
\newcommand{\LEGGP}{\mathrm{LEG}}
\newcommand{\PEG}{\mathrm{PEG}}

\newcommand{\PEGGP}{\mathrm{PEG}}

\DeclareMathOperator{\CyclicReduction}{CyclicReduction}
\DeclareMathOperator{\chol}{Cholesky}

\newcommand{\nobs}{n}
\newcommand{\nnotusdim}{d}
\newcommand{\nlat}{Q}
\newcommand{\ndim}{D}

\begin{document}
\maketitle
\begin{abstract}
Gaussian Processes (GPs) provide powerful probabilistic frameworks for interpolation, forecasting, and smoothing, but have been hampered by computational scaling issues.  Here we investigate data sampled on one dimension (e.g., a scalar or vector time series sampled at arbitrarily-spaced intervals), for which state-space models are popular due to their linearly-scaling computational costs.   It has long been conjectured that state-space models are general, able to approximate any one-dimensional GP.  We provide the first general proof of this conjecture, showing that any stationary GP on one dimension with vector-valued observations governed by a Lebesgue-integrable continuous kernel can be approximated to any desired precision using a specifically-chosen state-space model: the Latent Exponentially Generated (LEG) family. This new family offers several advantages compared to the general state-space model: it is always stable (no unbounded growth), the covariance can be computed in closed form, and its parameter space is unconstrained (allowing straightforward estimation via gradient descent).   The theorem's proof also draws connections to Spectral Mixture Kernels, providing insight about this popular family of kernels.  We develop parallelized algorithms for performing inference and learning in the LEG model, test the algorithm on real and synthetic data, and demonstrate scaling to datasets with billions of samples.
\end{abstract}

\section{Introduction}

Gaussian Process (GP) methods are a powerful and expressive class of nonparametric techniques for interpolation, forecasting, and smoothing \citep{GPbook}.  However, this expressiveness comes at a cost: if implemented naively, inference in a GP given $\nobs$ observed data points will require $O(\nobs^3)$ operations.  A large body of work has devised various means to circumvent this cubic run-time; briefly, this literature can be broken down into several threads.  A first approach is to attempt to perform exact inference without imposing any restrictions on the covariance kernel, using careful numerical methods typically including preconditioned conjugate gradients \citep{Cutajar2016}. \citet{Wang2019} represents the state of the art: inference and learning can be performed on $\sim 10^6$ datapoints on an 8-GPU machine and a few days of processing time.  A second approach searches for good approximations to the posterior (e.g., low-rank approximations to the posterior covariance) that do not rely on special properties of the covariance kernel.  Some well-known examples of this approach include   \citep{quinonero2005unifying,snelson2007local,hensman2013gaussian,low2015parallel,de2017gpflow}.   In a third approach, several techniques exploit special kernel structure.  Examples include matrices with Kronecker product structure \citep{gilboa2013scaling}, Toeplitz structure \citep{zhang2005time, cunningham2008}, {approximate bandedness \citep{gonzalez2009residual}}, matrices that can be well-approximated with hierarchical factorizations \citep{Ambikasaran2015},
or matrices which are sufficiently smooth to allow for interpolation-based approximations \citep{kiss-gp}.

When the GP has one-dimensional input, one popular method for scaling learning and inference is to approximate the GP with some form of Gaussian hidden Markov model (that is, a state-space model) \citep{reinsel2003elements,mergner2009applications,cheung2010estimation,brockwell2013time}.  This model class has considerable virtue: it is a particular case of a GP, it includes popular models like the auto-regressive moving average process (ARMA, when on an evenly spaced grid), and perhaps most importantly it admits linear-time $O(\nobs)$ inference via message passing.

It has long been conjectured that state-space models are not only fast, but general.  In several special cases it has been shown that state-space models provide arbitarily-good approximations of specific covariance kernels 
\citep{karvonen2016approximate,benavoli2016state,sarkka2019applied}.  Discrete-time processes on a finite interval can also be approximated this way \citep{lindgren2011explicit}.  The idea is simple: by taking a sufficiently high-dimensional latent space, one can produce processes whose observed covariance is quite intricate.  {Much progress has been made in understanding how quickly such approximations converge, i.e.\ how many latent dimensions are required \citep{hartikainen2010kalman,sarkka2014convergence,karvonen2016approximate}.} Practically, \citet{gilboa2013scaling} suggests it is straightforward to learn many GP models using state-space techniques.

However, no general proof of this conjecture exists, to our knowledge. 
\citet{karvonen2016approximate} provide a good example of the results in this vein that have already been achieved: theoretical tools for proving that particular Gaussian Process models can be approximated by state-space models.  These tools represent a promising advance from what came before, insofar as they can be applied to arbitrary kernels, allowing one to transform problems of kernel approximation into more elegant questions about Taylor series convergence.  However, they do have some disadvantages.  First, the proofs only show convergence for a compact subset of the Fourier transform of the kernels; thus the approximations might have arbitrarily poor high-frequency behavior.  Second, the methods do not apply if one measures a vector of observations at each time-point.

Here we present a new theorem which lays the ``generality'' question to rest for time-series data with vector-valued observations.  It is based on a connection between state-space models and Spectral Mixture kernels \citep{wilson2013gaussian}.   The new theorem shows that any stationary GP on one dimension with vector-valued observations and a Lebesgue-integrable continuous kernel can be uniformly approximated by a state-space model.   Along the way, we also prove a similar generality result for Spectral Mixture kernels of the form $\Sigma:\ \mathbb{R} \rightarrow\mathbb{R}^{\ndim\times \ndim}$.

To prove this theorem, we develop a convenient family of Gaussian hidden Markov models on one dimension: the Latent Exponentially Generated (LEG) process.  This model has several advantages over the general state-space models: its covariance can be computed in closed form; it is stable (with no unbounded growth); and its parameter space is unconstrained (allowing for simple gradient descent methods to be applied to learn the model parameters).   The LEG model generalizes the Celerit\'e family of Gaussian Processes \citep{foreman2017celerite} to allow more model flexibility and permit vector-valued observations.   Unlike some popular state-space models such as the ARMA, LEG 
processes do not require that the observations are equally spaced.  LEG kernels have an intersection with spectral mixture kernels \citep{wilson2013gaussian}, which enables a proof that these LEG kernels can approximate any integrable continuous kernel.  
Thus LEG models are general: our main mathematical result is to prove that for any stationary Gaussian Process $x$ on one dimension with integrable continuous covariance, for any $\varepsilon$, the covariance of $x$ can be uniformly matched within $\varepsilon$ by a LEG covariance kernel.  Finally, inference for the LEG processes can be parallelized efficiently via a technique known as Cyclic Reduction \citep{sweet1974generalized}, leading to significant runtime improvements.  

To summarize, this paper contains three contributions.  First, it is shown for the first time that any vector-valued GP on one dimension with a stationary integrable kernel can be approximated to arbitrary accuracy by a state-space model.  Second, an unconstrained parameterization for state-space models is developed (the LEG family).  Finally, based on that parameterization, this paper introduces an open-source, parallel-hardware, linear-time package for learning, interpolating, and smoothing vector-valued continuous-time state-space models with irregularly-timed observations.

The remainder of this paper defines the LEG family, explores its connections to other families of kernels, derives its generality, and finally empirically backs up these claims across real and synthetic data.  In particular, we show that the LEG family enables inference on datasets with billions of samples with runtimes that scale in minutes, not days.

                     
\section{Latent Exponentially Generated (LEG) kernels}

Consider a zero-mean Gaussian Process on one dimension, i.e. a random function $\obsV: \reals \rightarrow \reals^\ndim$ such that for any finite collections of times $t_1,t_2\cdots t_\nobs$ the distribution of $(\obsV(t_1),\cdots \obsV(t_\nobs)) \in \mathbb{R}^{\ndim \times \nobs}$ is jointly a zero-mean Gaussian.  The covariance kernel of $\obsV$ is a matrix-valued function defined by 
$   
\Sigma(s,t) = \mathbb{E}[\obsV(s)\obsV(t)^\top] \in \reals^{\ndim \times \ndim}
$.
A process is said to be stationary if $\Sigma(s,t)=C(s-t)$ for some matrix-valued function $C$ and $\mathbb{E}[x(t)]$ is the same for all values of $t$.  In this case we write $\tau$ for the time-lag $s-t$, i.e $C=C(\tau)$.   For clarity of exposition, what follows assumes stationarity and zero mean; generalizations are discussed in Section \ref{sec:extensions}.  

Some stationary Gaussian Processes are also \emph{linear state-space models}.  For the purposes of this article, we will say a process is a state-space model if there is a Markovian Gaussian Process $\latV:\ \mathbb{R}\rightarrow \reals^{\nlat}$ and at each timepoint $t$ we independently sample $\obsV(t) \sim \mathcal{N}(B\latV(t),\Lambda \Lambda^\top)$ with some matrices $B,\Lambda$. The process $\latV$ is called the ``latent'' process: it is used to define a distribution on the observations, $\obsV$, but it may not actually refer to any observable property of the physical world.  The matrices $B,\Lambda$ are referred to as the ``observation model.''

This paper was born from the following question: can \emph{any} stationary covariance kernel be approximated to arbitrary accuracy by a state-space model?

To answer this question, we found it convenient to design a family of linear state-space models which we call the ``Latent Exponentially-Generated'' family.  This family has several convenient features which enabled simpler analysis: the family can be indexed with an unconstrained parameter space, every member of the family is stationary, and it is easy to compute marginal covariances.  To define this family, we will start by defining a latent Markovian GP, then we will define the observation model.   Taken together these objects will form the LEG family of state-space models.  

\begin{definition} Let $\latV(0) \sim \mathcal{N}(0,I)$, let $\brownianmotion$ denote a standard Brownian motion, let $\diffusion,\rotparm$ be any $\nlat\times \nlat$ matrices, and let $G=\diffusion \diffusion^\top + \rotparm-\rotparm^\top$.  Let $\latV$ satisfy
$
\latV(t) = \latV(0) + \int_0^t\left(-\frac{1}{2}G \latV(s)ds + N d\brownianmotion(s)\right).
$
Then we will say $\latV$ is a Purely Exponentially Generated process, $\latV \sim \PEGGP(\diffusion,\rotparm)$.  Applying the standard formulae for the covariances of linear stochastic differential equations, we obtain that the covariance kernel of $\latV$ is given by the formula below (we refer the reader to Appendix \ref{sec:ap_known} for a brief review).
\[
C_{\PEG}(\tau;\diffusion,\rotparm)  = \begin{cases}
\exp\left(-\frac{|\tau|}{2}G\right) & \mathrm{if}\ \tau\geq0 \\ 
\exp\left(-\frac{|\tau|}{2}G^\top \right) & \mathrm{if}\ \tau\leq0
\end{cases}
\] 
If the real parts of the eigenvalues of $G$ are strictly positive,  the spectral representation of the PEG kernel may be calculated as 
\[
C_\mathrm{PEG}(\tau;N,R) = \frac{1}{2\pi} \int \exp(-i\omega \tau) \left[(G/2 - i\omega I)^{-1} + (G^\top/2 + i\omega I)^{-1} \right]
d \omega
\]
We refer the reader to \cite{cramer1940theory} for a collection of first properties on such spectral representations; we also briefly review these properties in Appendix \ref{sec:ap_known}.
\end{definition}

\begin{figure*}
\vskip 0.2in
\begin{center}
\centerline{\includegraphics[width=.9\textwidth]{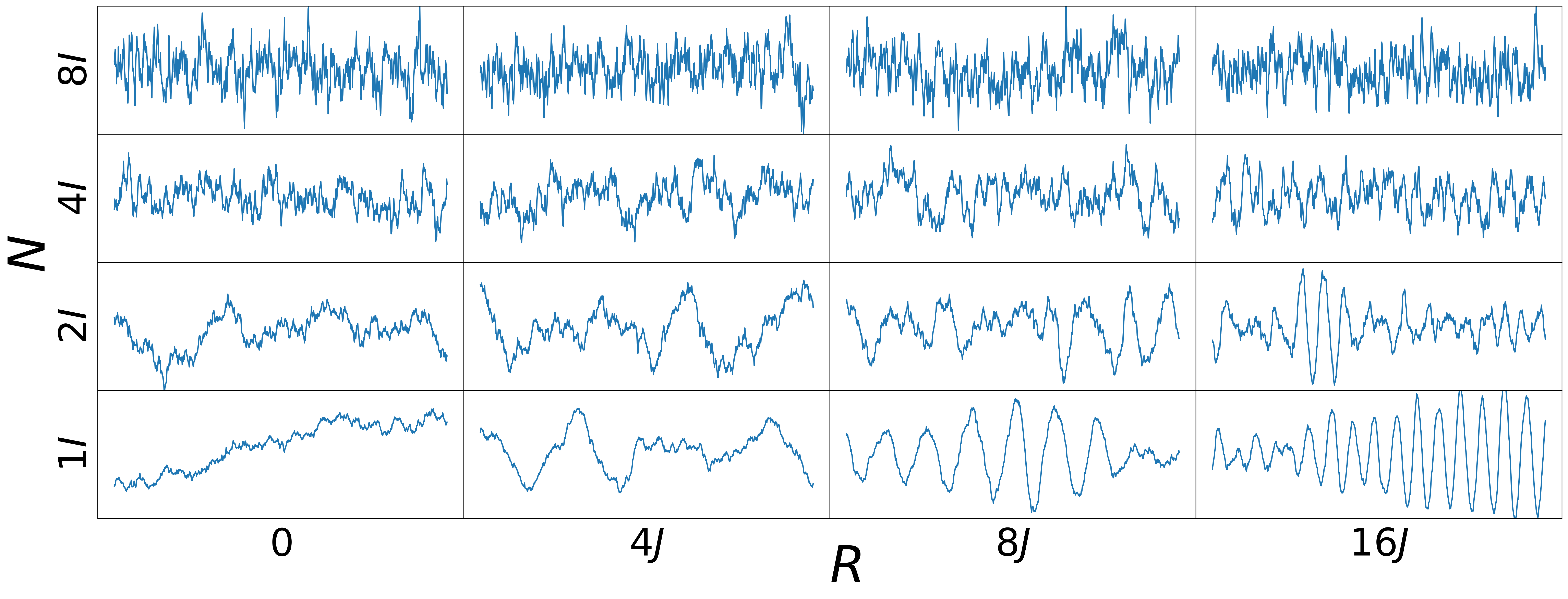}}
\caption{\textbf{PEG process samples}.  The plots above show representative samples from the model $\PEG(\diffusion,\rotparm)$ as we vary $\diffusion$ and $\rotparm$.  Here we consider rank-2 PEG models (only one element of the 2d vector is plotted), so $\diffusion,\rotparm$ are both $2\times 2$ matrices.   We vary $\diffusion$ by taking it to be various multiples of the identity.  We vary $\rotparm$ by taking various multiples of $J$, the antisymmetric $2\times 2$ matrix with zeros on the diagonal and $\pm 1$ on the off-diagonal.  In this simple rank-2 case, increasing $N$ leads to a less predictable process and  increasing $\rotparm$ leads to faster oscillations.
\label{fig:exampleprocs}}
\end{center}
\vskip -0.2in
\end{figure*}
 
The matrices $\diffusion,\rotparm$ can be interpreted intuitively.  The positive definite diffusion $\diffusion \diffusion^\top$ controls the predictability of the process: when an eigenvalue of $\diffusion \diffusion^\top$ becomes larger, the process $\latV$ becomes less predictable along the direction of the corresponding eigenvector.  The antisymmetric  $\rotparm-\rotparm^\top$ term affects the process by applying an infinitesimal deterministic rotation at each point in time.  The eigenvalues of $\rotparm - \rotparm^\top$ are purely imaginary, and when they are large they lead to strong rapid oscillations in the process, while the eigenvectors control how these oscillations are mixed across the dimensions of $\latV$.  As an illustration, Figure \ref{fig:exampleprocs} shows the first dimension of samples from a $\nlat=2$ PEG process with various values of $\diffusion,\rotparm$. 

We now turn to the observed process: 

\begin{definition}
Let $\latV \sim \PEGGP(\diffusion,\rotparm)$.  Fix any $\ndim \times \nlat$ matrix $B$ and any $\ndim \times \ndim$ matrix $\Lambda$.   For each $t$ independently, define the conditional observation model: 
$
\obsV(t)|\latV(t) \sim \mathcal{N}(B \latV(t),\Lambda \Lambda ^\top).
$
We define a \textbf{Latent Exponentially Generated (LEG)} process to be the Gaussian Process $\obsV: \reals \rightarrow \reals^\ndim$ generated by a PEG prior and the above observation model.  We write $\obsV \sim \LEGGP(\diffusion,\rotparm,B,\Lambda)$.  The formula for the covariance of $\obsV$ is given below.
\[
C_\LEG(\tau;\diffusion,\rotparm,B,\Lambda) = \begin{cases}
B \left(C_\PEG(\tau;\diffusion,\rotparm)\right) B^\top & \mathrm{if}\ \tau=0 \\
B \left(C_\PEG(\tau;\diffusion,\rotparm)\right) B^\top + \Lambda \Lambda^\top  & \mathrm{otherwise}
\end{cases}
\]
When $\Lambda=0$, we may suppress that parameter by the abuse of notation $C_\mathrm{LEG}(N,R,B)=C_\mathrm{LEG}(N,R,B,0)$.  We will refer to the latent dimension $\nlat$ as the \textbf{rank} of the LEG kernel and $\ndim$ as the \textbf{dimension} of the kernel.  

\end{definition}

In the next section we will see how the LEG family of kernels is connected to other popular families.  Before we do so, let us collect a few elementary properties about the LEG family itself.

\begin{lemma}[Connections among LEG kernels]  \label{lem:legclosure} Let $\Sigma=C_\mathrm{LEG}(N,R,B,\Lambda)$ denote a LEG kernel of dimension $\ndim$ and rank $\nlat$.  Let $\tilde \Sigma =C_\mathrm{LEG}(\tilde N,\tilde R,\tilde B,\tilde \Lambda)$ denote a LEG kernel of dimension $\tilde \ndim$ and rank $\tilde \nlat$.   Let $\otimes$ denote the Kronecker product and $\oplus$ denote the direct sum.
 
\begin{enumerate}
    \item If $\ndim = \tilde \ndim$, then the kernel $\tau \mapsto \Sigma(\tau)+\tilde \Sigma(\tau)$ lies in the family of LEG kernels of rank $\nlat+\tilde \nlat$.
    \item The kernel $\tau \mapsto \Sigma(\tau) \otimes \tilde \Sigma(\tau)$ lies in the family of LEG kernels of rank $\nlat\tilde \nlat$ and dimension $\ndim \tilde \ndim$.
    \item Let $\gamma>0$.  The kernel $\tau \mapsto \Sigma(\tau/\gamma)$ lies in the family of LEG kernels of rank $\nlat$ and dimension $\ndim$.
\end{enumerate}

\end{lemma}

\begin{proof} We can construct each new kernel explicitly in the LEG form.
\begin{enumerate}
    \item Take $C_\mathrm{LEG}(N\oplus \tilde N,R \oplus \tilde R,(B\  \tilde B), \Lambda + \tilde \Lambda)$.
    \item Take $C_\mathrm{LEG}((NN^\top\otimes I + I\otimes \tilde N \tilde N^\top)^{1/2},R\otimes I + I\otimes \tilde R, B \otimes \tilde B, (\Lambda\Lambda^\top \otimes \tilde \Lambda \tilde \Lambda^\top)^{1/2})$.  To see that this kernel matches the desired Kronecker product, we use the fact that $\exp(A\otimes I + I\otimes \tilde A)=\exp(A)\otimes\exp(\tilde A)$; we refer the reader to \citet{neudecker1969note} for exposition on this subject.
    \item Take $C_\mathrm{LEG}(N/\sqrt{\gamma},R/\gamma,B,\Lambda)$.
\end{enumerate}
\end{proof}


\section{Connections between LEG kernels and other model families}

We here summarize the relationship between the LEG family and several popular families of stationary positive-definite kernels on one dimension.   We first define each family in turn and then present a Lemma describing some connections between these families and the LEG family.

\begin{description}
    \item[Spectral mixtures.] Spectral mixture kernels, as introduced by \citet{wilson2013gaussian}, define a covariance kernel by taking the Fourier transform of a weighted sum of different shifts of a probability density.  It was originally designed for GPs of the form $\obsV:\ \mathbb{R}^\nnotusdim \rightarrow \reals$; {here we generalize this to ones of the form $\obsV:\ \reals^\nnotusdim \rightarrow \reals^\ndim$}.  Let $\hat K:\ \mathbb{R}^\nnotusdim \rightarrow \mathbb{R}$ denote any positive definite kernel.  Let $\gamma>0$, and $b \in \mathbb{C}^\ndim$.  We define a Spectral Mixture Term by \[
    C_\mathrm{SMT}(\tau;\hat K,b,\mu,\gamma) = \Re(\exp(-i\tau \mu)b b^*)\hat K(\tau/\gamma),
    \]
    where $\Re(x)$ takes the real part of a matrix and $b^*$ denotes the conjugate transpose of $b$.  We will say such a term is \textbf{based on $\hat K$}.  Spectral mixture terms are the result of applying six transformations to $\hat K$: taking the Fourier transform, scaling the domain by $\gamma$, shifting the domain by $\mu$,  multiplying by $bb^*/\gamma$, taking the inverse Fourier transform, and taking the real part.  We define a Spectral Mixture kernel $C_\mathrm{SM}:\ \mathbb{R}^\nnotusdim \rightarrow \mathbb{R}^\ndim$ with $\nlat$ components to be any sum of $\nlat$ spectral mixture terms each carrying the same choice for $\hat K,\gamma$.  
    
    \item[Matern.] Let $K_\nu$ denote the modified bessel function of the second kind.  The standard Matern kernel of order $\nu$ is given by $C_\mathrm{MTN}(\tau;\nu)=2^{1-\nu}(\tau\sqrt{2\nu})^\nu K_\nu(\tau \sqrt{2\nu})/\Gamma(\nu)$.  We refer the reader to \cite{sarkka2019applied} for details on this popular family of kernels.  
    
    \item[Celerit\'e terms.] Developed by  \cite{foreman2017celerite}, a Celerit\'e term is given by $C_\mathrm{CLR}(\tau;a,b,c,d)= a\exp(-c |\tau|) \cos(d|\tau|) + b\exp(-c|\tau|) \sin(d|\tau|)$.  Such terms are not always valid kernels, i.e. there are parameters $a,b,c,d$ such that $C(\cdot;a,b,c,d)$ is not the kernel of \emph{any} Gaussian Process.  If a Celerit\'e term is the kernel of a Gaussian Process, it is said to be ``positive-definite.''
    
    \item[Stationary Ornstein-Uhlenbeck.]  Let $G\in \reals^{\nlat \times \nlat}$ such that that the real part of the eigenvalues of $G$ are strictly positive.  Let $N \in \reals^{\nlat \times \nlat}$ denote any other matrix.  Then there exists a unique symmetric nonnegative-definite matrix $P_\infty$ satisfying the continuous Lyapunov equation $GP_\infty + P_\infty G^\top = 2NN^\top$ (cf. Chapter 2 of \citet{artemiev1997numerical}).  Let $\latV(0)\sim \mathcal{N}(0,P_\infty)$ and
    $
    \latV(t) = \latV(0) + \int_0^t\left(-\frac{1}{2}G \latV(s)ds + N d\brownianmotion(s)\right)
    $,
    where $\brownianmotion$ is a standard Brownian motion. Then $\latV$ is a stationary Gaussian process, known as the Ornstein-Uhlenbeck process parameterized by $G,N$.  The covariance kernel of $\latV$ is given below (we refer the reader to Appendix \ref{sec:ap_known} for a brief review of such covariances).
    \[
    C_\mathrm{OU}(\tau;G,N) = \begin{cases}
        \exp(-G|\tau|/2)P_\infty & \mathrm{if}\ \tau\geq 0 \\
        P_\infty\exp(-G^\top|\tau|/2)  & \mathrm{if}\ \tau\leq 0 
    \end{cases}
    \]
    We will refer to $\nlat$ as the \textbf{dimension} of the kernel.  Note that if $G+G^\top=2NN^\top$ then $P_\infty=I$ and this is exactly equal to a PEG kernel.
    
    \item[Stationary linear state-space.] Let $C_\mathrm{OU}(G,N)$ denote a stationary Ornstein-Uhlenbeck kernel of dimension $\nlat$.  Fix any matrix $B \in \reals^{\ndim \times \nlat}$.  We define the stationary state-space model kernel family as follows:
    \[
    C_\mathrm{SSM}(\tau;G,N,B) = B C_\mathrm{OU}(\tau; G,N) B^\top.
    \]
    We will refer to $\nlat$ as the \textbf{rank} of the kernel and $\ndim$ as the \textbf{dimension} of the kernel.  {Such kernels are always nonnegative definite, though they are not guaranteed to be strictly positive definite; for example, if $\nlat < \ndim$ the kernel will fail to be strictly positive definite.} 
\end{description}

Below we summmarize a few connections between these families and LEG kernels.

\begin{lemma}[Connections to other model families]    \label{lem:smareleg}
\hspace{.01in}

\begin{enumerate}
    \item Let $G \in \reals^{\nlat \times \nlat}$ such that the real part of all eigenvalues is strictly positive.  Let $N\in \reals^{\nlat \times \nlat},B\in \reals^{\ndim \times \ndim}$.  Then  $C_\mathrm{SSM}(G,N,B)$ lies in the family of LEG kernels of rank $\nlat$.
    \item Let $\nu \in \{1/2,3/2,\cdots \}$.  Then the Matern kernel of order $\nu$ lies in the family of LEG kernels of rank $\nu+1/2$.
    \item Let $C_\mathrm{LEG}(N,R,B)$ denote a LEG kernel of dimension 1 and rank $\nlat$.  Then every spectral mixture kernel with $\tilde \nlat$ components based on $C_\mathrm{LEG}(N,R,B)$ lies in the family of LEG kernels with rank $2\tilde \nlat \nlat$. 
    \item Every positive-definite Celerit\'e term lies in the family of LEG kernels with rank 2.
\end{enumerate}
\end{lemma}

\begin{proof} We treat each point in turn.
\begin{enumerate}
    \item Let $P_\infty = C_\mathrm{OU}(0; G,N)$.  Without loss of generality, we assume $P_\infty$ is strictly positive definite (if any eigenvalue of $P_\infty$ is zero, it follows that $\alpha^\top \latV(t)=0$ for all $t$ for the corresponding eigenvector $\alpha$; we may then adopt a change of variables to a new family of state-space processes without this degeneracy).  Fix any invertible matrix $\Psi$ such that $\Psi P_\infty \Psi^\top = I$ (for example, the principal square root of the inverse of $P_\infty$ will always suffice).  Let $\tilde G=\Psi G \Psi^{-1}$, $\tilde N=\Psi N$, $\tilde B= \Psi^{-1}B$, and $R= (\tilde G - \tilde G^\top) /4$.  The continuous Lyapunov equation $GP_\infty + P_\infty G^\top = 2NN^\top$ yields that  $C_\mathrm{SSM}(\tau;G,N,B) = C_\mathrm{LEG}(\tau;\tilde N,R,\tilde B)$.
    
    \item Combine point 1 of this lemma with the corresponding result for state-space models (cf. Chapter 12 of \citet{sarkka2019applied}).
    
    \item  Let $\mu\geq 0$ and $b\in \mathbb{C}^\ndim$.  In light of Lemma \ref{lem:legclosure}, it suffices to show that $\Re(\exp(-i\tau\mu)bb^*)$ lies in the LEG family.  We start by defining a skew-symmetric matrix $J$, below.
    \[
    J=\left(\begin{array}{cc}
    0 &  1\\
    -1 & 0
    \end{array}\right)
    \]
    Let $\Sigma(\tau)=C_\mathrm{LEG}(\tau;0,\mu J,(\Re(b) \  \Im(b)))$.  We calculate.
    \[
    \Sigma(\tau)=
        \cos(\mu\tau)\left(\Re(b)\Re(b)^{\top}+\Im(b)\Im(b)^{\top}\right)
        +\sin(\mu\tau)\left(-\Re(b)\Im(b)^{\top}+\Im(b)\Re(b)^{\top}\right)
    \]
    Applying Euler's formula, we obtain $\Sigma(\tau)= \Re(\exp(-i \tau \mu)bb^*)$, as desired. 
    
    \item Let $N_1 = \sqrt{2c-2bd/a}$, $R_1=\sqrt{2c^2 + 4d^2 + 2b^2 d^2/a^2}$ and $N_2=\sqrt{c+bd/a}$.  \citet{foreman2017celerite} show that that $|bd|<ac$ and $a,c\geq 0$ for any positive-definite Celerit\'e term, thus $N_1,R_1,N_2 \in \mathbb{R}$.  Let 
    \[
    N= \left(\begin{array}{cc}
    N_1 & 0\\
    N_2 & N_2
    \end{array}\right)\quad R=\left(\begin{array}{cc}
    0 &  R_1\\
    0 & 0
    \end{array}\right)\quad B=\left(\begin{array}{cc}
    \sqrt{a} & 0\end{array}\right)
    \]
    Then observe that $C_\mathrm{CLR}(a,b,c,d)=C_\mathrm{LEG}(N,R,B)$.
\end{enumerate}
\end{proof}

\section{Generality of LEG kernels and Spectral Mixtures for vector-valued time-series}

Here we show that spectral mixture and LEG kernels are dense in the uniform norm in the space of continuous integrable positive definite kernels on one dimension.  We additionally show that for each $\xi \in \mathbb{N}$, the subset of LEG kernels associated with $\xi$-times-differentiable processes is also dense in this space.  

To obtain these results we develop a generalization of a classic theorem from the kernel density estimation literature.

\begin{theorem} \label{lem:kdereg}
Let $K,p$ denote bounded densities on $\mathbb{R}^d$.  Let $g:\ \mathbb{R}^d \rightarrow [-M,M]^D$.  Let $\gamma_\ell = \ell^{1/2d}$.  Let $\mu_1,\mu_2 \cdots \sim p$, independently.  For each $\ell \in 1,2,\cdots$, define  
$
h_{\ell}(\xi) = \frac{1}{\ell} \sum_{k=1}^{\ell}g(\mu_k) \gamma^d_\ell K(\gamma_\ell(\xi - \mu_k))
$.  
Then
\[
\mathbb{P}\left(\lim_{\ell \rightarrow \infty} \sum_{i=1}^D \int |h_{\ell,i}(\xi) - p(\xi)g_i(\xi)| d\xi  = 0\right) = 1.
\]
\end{theorem}
\begin{proof} We largely imitate the proof of Devroye and Wagner \citep{devroye1979l1}, which handles the special case that $g(x)= 1$.  

Let us start by assuming $g$ takes the form $g:\ \mathbb{R}^d \rightarrow [0,M]$. 

For almost any fixed $\omega$, we have that $\lim_{\ell \rightarrow \infty}|h_\ell(\omega) - p(\omega) g(\omega)|=0$  almost surely.  This follows from two steps:

\begin{enumerate}
\item \emph{Controlling the bias}.  Let 
\begin{align*}
\bar h_{\ell}(\omega)
&= \mathbb{E}\left[h_{\ell}(\omega)\right]\\
&= \int g(x) \gamma_\ell^d K(\gamma_\ell(\omega - x)) p(x)dx
\end{align*}

Now fix any $\delta>0$.  We have that 
\begin{align*}
|\bar h_{\ell}(\omega) - p(\omega)g(\omega)| 
&\leq \int_{\Vert x-\omega \Vert < \delta/\gamma_\ell} |p(x)g(x) - p(\omega)g(\omega)| \gamma^d_\ell K(\gamma_\ell(\omega - x)) dx\\
&\qquad +  \int_{\Vert x-\omega \Vert \geq \delta/\gamma_\ell} |p(x)g(x) - p(\omega)g(\omega)| \gamma^d_\ell K(\gamma_\ell(\omega - x)) dx\\ 
\end{align*}
We look at each term separately:
\begin{itemize}
    \item For $x \approx \omega$ we apply the Lebesgue differentiation theorem.  Let $c=\sup K(x)$ and let $\lambda(\delta)$ denote the volume of the ball of radius $\delta$.  Noting that $\gamma^d_\ell = \lambda(\delta)/\lambda(\delta/\gamma_\ell)$, we see that the integral of the error over $\Vert x-\omega \Vert < \delta/\gamma_\ell$ is bounded by
    \begin{align*}
         c \lambda(\delta) \frac{1}{\lambda(\delta/\gamma_\ell)} \int_{\Vert x-\omega \Vert < \delta/\gamma_\ell} |p(x)g(x) - p(\omega)g(\omega)| dx
    \end{align*}
    For any fixed $\delta$, the Lebesgue differentiation theorem shows that this goes to zero almost everywhere because $\gamma \rightarrow \infty$.
    
    \item For $\Vert x-\omega\Vert > \delta/\gamma_\ell$.  Let $c=M\sup_x p(x)$.  Then the integral of the error over this domain is bounded by
    \[
    2c\int_{\Vert x-\omega \Vert \geq \delta} K(\omega - x) dx
    \]
    Note that we used a change of variables to drop any dependency on $\gamma_\ell$. Since $K$ is a density we can always find $\delta$ so that this is arbitrarily small.
\end{itemize}

Therefore, for any fixed $\varepsilon$ we can always find a $\delta$ which ensures that the second term is less than $\varepsilon/2$, and then ensure that the first term is less than $\varepsilon/2$ for all sufficiently large $\ell$.  In short, $|\bar h_{\ell}(\omega) - p(\omega)g(\omega))| \rightarrow 0$ for each $\omega$.

\item \emph{Controlling the variation}.  Now we would like to bound $\bar h_{\ell}(\omega) -  h_{\ell}(\omega)$.  To do this we note that it is a sum of independent random variables of the form $g(\mu_k)\gamma_\ell^d K(\gamma_\ell (\omega-\mu_k))/\ell$.  Letting $c=M \sup_x K(x)$ we observe that the absolute value of each random variable is bounded by $\gamma^d c/\ell$.  Hoeffding's inequality then gives that 
\begin{gather*}
\mathbb{P}\left(\left|\bar h_{\ell}(\omega) -  h_{\ell}(\omega)\right| > \varepsilon\right) 
\leq 2\exp\left(-\frac{2\ell t^2}{\gamma^d_\ell c}\right)
= 2\exp\left(-\sqrt{\ell} \frac{2 t^2}{c}\right)
\end{gather*}
The right-hand-side is always summable for any $t>0$.  Indeed, one may readily verify that if $f(x)=-2\exp(-c\sqrt{x})(c\sqrt{x}+1)/c^2$, then $f'(x)=\exp(-c\sqrt{x})$.  For any $c>0$ it follows that $\int_1^\infty \exp\left(-c\sqrt{x}\right) dx=2(c+1)e^{-c}/c^2 < \infty$ and 
\[
\sum_\ell \mathbb{P}\left(\left|\bar h_{\ell}(\omega) -  h_{\ell}(\omega)\right| > \varepsilon\right) < \infty 
\]
Applying Borel-Cantelli we find that $|\bar h_{y,\ell}(\omega)-h_{y,\ell}(\omega)|$ converges almost surely to zero.  
\end{enumerate}
Combining these steps together, we obtain a pointwise result: for almost every $\omega$, the sequence  $h_1(\omega),h_{2}(\omega),\cdots$ converges almost surely to $p(\omega)g(\omega)$.  

We must now extend this pointwise result to $\mathscr{L}^1$.  To do so we start by noting that $\int |h_\ell(\omega)|d\omega \rightarrow \int |p(\omega)g(\omega)|d\omega$ almost surely.  Since $g(\omega)\geq 0$, we have that 
\[
\int |h_\ell(\omega)|d\omega = \int h_\ell(\omega)d\omega = \frac{1}{\ell} \sum_{k=1}^{\ell} g(\mu_k)
\]
Note that this is the average of independent and identically distributed bounded objects.  The law of large numbers thus gives us that $\int |h_\ell(\omega)|d\omega$ converges almost surely to $\mathbb{E}_{\mu_1 \sim p}[g(\mu_1)]=\int p(\omega)g(\omega)d\omega=\int |p(\omega)g(\omega)|d\omega$.  This allows us to extend our pointwise result to the desired $\mathscr{L}^1$ result via Glick's Lemma, which is stated for the benefit of the reader in Lemma \ref{lem:glick} in Appendix \ref{sec:ap_known}.

Finally, let us consider the case of $g: \reals^d \rightarrow [-M,M]^D$.  For each $i \in \{1 \cdots D\}$ apply the above arguments twice (once for the positive part of $g_i$ and once for the negative part of $g_i$) together with the triangle inequality to yield a proof for full statement of this theorem. 
\end{proof}

With this in hand, the results we seek are straightforwad.

\begin{theorem} \label{thm:specawesome} Fix any positive-definite Lebesgue-integrable kernel {$\hat K:\ \reals^\nnotusdim \rightarrow \reals$} with $K(0)=1$, any $\varepsilon>0$, and any continuous Lebesgue-integrable covariance kernel of the form $\Sigma: \mathbb{R}^\nnotusdim \rightarrow \mathbb{R}^{\ndim \times \ndim}$.  One may find a spectral mixture kernel based on $\hat K$ which uniformly approximates $\Sigma$ to within $\varepsilon$.
\end{theorem}

\begin{proof}
    Apply Bochner's theorem to find the bounded density $K$ such that $\hat K(\tau)=\int\exp(-i \tau\cdot \omega)K(\tau)d\tau$.  Apply Bochner's theorem again to find a bounded density $f$ and bounded Hermitian nonnegative-definite matrix-valued function $M$ such that
    \[
    \Sigma(\tau) = \int \exp(i \tau \cdot \omega) f(\omega) M(\omega) d\omega
    \]
    We refer the reader to Appendix \ref{sec:ap_known} for a review of these matrix-valued spectra.  Apply a unitary eigendecomposition to each $M(\omega)$, obtain $b_i(\omega)$ such that $M(\omega)=\sum_i^{\ndim}b_i(\omega)b_i^*(\omega)$. Sample $\mu_1,\mu_2,\mu_3,\cdots$ independently and identically from $f$.  For any fixed value of $\nlat$, we can select a value for $\gamma$ and define a mixture of $\nlat \ndim$ components as follows.
    \[
    \hat M(\omega;\nlat) = \frac{1}{\nlat} \sum_i^\ndim \sum_k^\nlat 
        b_i(\mu_k) b_i^*(\mu_k) \gamma_\nlat K(\gamma_\nlat(\omega - \mu_k))
    \]
    Applying Theorem \ref{lem:kdereg}, we may choose $\gamma_\nlat$ such that $\hat M$  almost surely converges in total variation to $fM$ as $\nlat \rightarrow \infty$.    Let $C(\tau)=\int\exp(-i\tau \omega)M(\omega)d\omega$.  The total variation convergence of $\hat M$ yields uniform convergence of $C$ to $\Sigma$.  One can thus achieve any degree of accuracy by increasing $\nlat$ until the uniform error drops below the desired level.  Finally, to obtain a (real-valued) spectral mixture kernel which uniformly approximates $\Sigma$, take the real part of $C$.
\end{proof}

{The theorem above covers kernels of the form $\mathbb{R}^\nnotusdim \rightarrow \mathbb{R}^{\ndim \times \ndim}$, though our focus here is on kernels of the form $\mathbb{R} \rightarrow \mathbb{R}^{\ndim \times \ndim}$. In particular, combining Theorem \ref{thm:specawesome} with Lemma \ref{lem:smareleg} we obtain the main result we sought.  }

\begin{theorem}[Flexibility of LEG kernels] \label{thm:legflex}
For every $\xi \in \mathbb{N},\varepsilon>0$ and every Lebesgue-integrable continuous positive definite stationary kernel $\Sigma: \mathbb{R} \rightarrow \mathbb{R}^{n\times n}$ there exists a LEG kernel $C$ which is associated with a $\xi$-times differentiable process and differs from $\Sigma$ by at most $\varepsilon$.
\end{theorem}

\begin{proof}
Use Theorem \ref{thm:specawesome} to obtain a spectral mixture kernel, based on the Matern kernel of order $\xi+1/2$ and achieving the desired accuracy.  Recall that processes with such Matern-based kernels are $\xi$-times-differentiable.  Finally, use Lemma \ref{lem:smareleg} to obtain a LEG representation of this spectral mixture.     
\end{proof}

\section{Black-box Gaussian Processes on one dimension}

\label{sec:leggpcomputation}

The theoretical results above suggest it may be possible to produce a black-box solution for efficient parameter estimation, interpolation/extrapolation, and smoothing for Gaussian Processes on one dimension.  That is, it should be possible to design algorithms which can robustly perform all of these tasks without requiring a user to specify anything more than the rank $\nlat$ of the LEG kernel to be learned.  

To test this hypothesis, we developed a collection of algorithms to enact such a black-box solution.  We implemented these algorithms in the pure-python \texttt{leggps} package (\url{https://github.com/jacksonloper/leg-gps}).  We sketch the main ideas below.

\begin{itemize}
    \item Perform parameter estimation by gradient descent on the likelihood.  This is straightforward to implement because the space of LEG parameters is unconstrained.  Although the likelihood is not convex, we found in our experiments (see below) that random initialization was sufficient to find good parameter estimates. 
    
    \item Use dynamic programming to enable linear-time computations for likelihood, interpolation/extrapolation, and smoothing.  Dynamic programming algorithms are the standard tool for analyzing state-space models \citep{li2009markov}.  When properly designed, the total computational cost of such algorithms scales linearly with the number of observations.  
    \item Use the Cyclic Reduction (CR) variable-ordering for this dynamic programming, in order to obtain a parallelizable algorithm.  Recall that any dynamic programming algorithm requires the specification of the order in which variables are processed.  If one uses the time-points to order the variables (processing them one at a time along the state-space chain) one obtains the popular Kalman Filter algorithm.  Unfortunately, this algorithm cannot be easily parallelized to take advantage of modern hardware such as GPUs.  The CR ordering, well-known in the matrix algorithms literature \citep{sweet1974generalized}, yields parallelizable algorithms with state-space models.  The CR ordering proceeds in roughly $\log_2 m$ stages.  At each stage all the relevant dynamic programming computations can be performed entirely in parallel.  The first stage handles every other variable (e.g. all the variables with even-valued indices).  The second stage then applies the same idea recursively, processing every other variable of that which remains.  For example, given 14 variables, the CR ordering proceeds in 5 stages, given by
    \[
    \overbrace{2,4,6,8,10,12,14}^1,\overbrace{3,7,11}^2,\overbrace{5,13}^3,\overbrace{9}^4,\overbrace{1}^5.
    \]
    This ordering yields parallelizable dynamic programming algorithms for all of the matrix operations necessary for computation of likelihoods, interpolation/extrapolation, and smoothing.  Details can be found in Appendix \ref{sec:ap_known}, but here we summarize the key operations.  Let $J$ denote a block tridiagonal matrix with $\nobs$ blocks of size $\nlat$.   We need to be able to compute the determinant of $J$, solve linear systems $Jx=y$, compute $y^\top J^{-1}y$ (this can in fact be done in half the time it takes to solve $Jx=y$), and compute the diagonal and off-diagonal blocks of $J^{-1}$.  Given $k<\nobs$ processors, each of these algorithms can be completed in $O(\nlat^3\nobs / k)$ clock cycles using dynamic programming with a CR ordering.  
    
    CR has similarities to other techniques.  It is intuitively similar to multigrid techniques \citep{terzopoulos1986image,hackbusch2013multi} which have been used for Gaussian inference in other contexts \citep{papandreou2010gaussian,mukadam2016gaussian,zanella2017analysis}, but CR is a direct exact method whereas multigrid is iterative.  CR is quite similar to the associative-scan method developed by \citet{sarkka2019temporal}.  Like CR, associative scans use parallel hardware to get exact calculations for state-space models.   In practice, we used CR because we found it easy to write the code to calculate all the relevant quantities (e.g. posterior variances).   Implementing CR in modern software libraries (TensorFlow2 in this case) was straightforward, making it easy to take advantage of modern hardware.  
    
    \item Compute the gradients of matrix exponentials efficiently using the work of \cite{najfeld1995derivatives}.  Computation of $\exp(-G\tau/2)$ for many different values of $\tau$ can be achieved quickly using an eigendecomposition, but the same idea cannot be directly applied to compute the gradient of $\exp(-G\tau/2)$ with respect to $G$.  Fortunately, \cite{najfeld1995derivatives} derive an explicit expression for this gradient in terms of the eigendecomposition of $G$.  Details can be found in Appendix \ref{sec:apdxgrad}.
\end{itemize}


\section{Extensions}

\label{sec:extensions}

Before moving on to illustrate these results with experiments on simulated and real data, we pause to note several useful extensions:

\emph{Nonstationary processes}. We have focused on stationary processes here for simplicity.  A number of potential extensions to nonstationary processes are possible while retaining linear-time scaling.  For example, time-warped versions of a stationary processes could be accurately modeled by a time-warped LEG model, assuming the time-warping is Lipschitz-continuous.   \citet{benavoli2016state} suggests another direction for nonstationarity processes with state-space models, achieved by introducing determinism into the relevant stochastic differential equations.  Finally, switching linear dynamical systems and the recurrent versions thereof offer a promising direction for moving beyond the stationary Gaussian model \citep{linderman2017bayesian}; the ideas presented here fit without alteration into those frameworks.  

\emph{Non-Gaussian observations}.  Many approaches have been developed to adapt GP inference methods to non-Gaussian observations, including Laplace approximations, expectation propagation, variational inference, and a variety of specialized Monte Carlo methods \citep{hartikainen2011sparse,riihimaki2014laplace,nguyen2014automated,nishihara2014parallel}.  Many of these can be easily adapted to the LEG model, using the fact that the sum of a block-tridiagonal matrix (from the precision matrix of the LEG prior evaluated at the sampled data points) plus a diagonal matrix (contributed by the likelihood term of each observed data point) is again block-tridiagonal, leading to linear-time updates \citep{smith2003estimating,Paninski2010,fahrmeir2013multivariate,polson2013bayesian,Khan,Nickisch}.

\emph{Tree-structured domains}.  Just as Gaussian Markov models in discrete time can be easily extended to Gaussian graphical models on tree-structured graphs, it is straightforward to extend the Gaussian Markov PEG and LEG processes to stochastic processes on more general domains with a tree topology, where message-passing can still be applied.  

\emph{{Multidimensional domains}}.
We may also be able to adapt state-space models for multi-dimensional Gaussian processes of the form $x:\ \mathbb{R}^d\rightarrow\mathbb{R}^n$, with $d>1$.  Given collections of matrices $N,R,B$, consider the ``sum-of-separable terms" kernel $\tau \mapsto \sum_{r=1}^{\zeta} \prod_{k=1}^d C_\LEG(\tau_k;N_{rk},R_{rk};B_{rk}))$ where $\prod$ denotes a Hadamard product.     This family of kernels has a number of nice properties.  First, the Schur product theorem readily shows that such kernels are nonnegative definite.   Second, this family of kernels includes spectral mixtures of the form required by Theorem \ref{thm:specawesome} (so it is dense among stationary continuous Lebesgue-integrable kernels of the form $\Sigma:\ \mathbb{R}^d\rightarrow \mathbb{R}^{n \times n}$).  Finally, as shown in Appendix \ref{sec:mmkronhad}, efficient computation with such kernels is possible if our observations all lie within a (potentially irregularly-spaced) grid: the cost of matrix-multiplication by the covariance will scale linearly with the number of points in the grid.  Preconditioned conjugate gradient methods such as those implemented in GPyTorch (cf.\ \citet{gardner2018gpytorch}) require nothing more than efficient covariance matrix-mutliplications; this kind of approach could thus yield efficient inference algorithms for such processes.


\section{Experiments} 

\subsection{Model rank versus approximation error}

\begin{figure*}
\vskip 0.2in
\begin{center}
\centerline{\includegraphics[width=.9\textwidth]{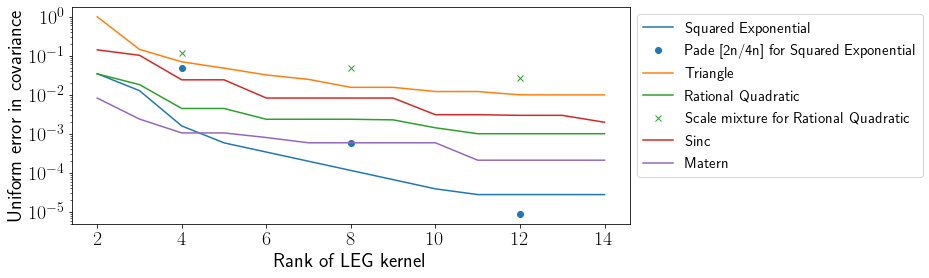}}
\caption{\textbf{Popular kernels can be approximated to within 1\% of their maximum value using a rank 7 LEG model}.  We used gradient descent to find LEG kernels of various ranks to match a few popular one-dimensional kernels.  We also examined two proposals from the literature for approximating these kernels with state-space models (the Pade [2n/4n] proposal for squared exponential kernels and the scale mixture proposal for rational quadratic kernels); in most cases we found that gradient descent was able to find a better state-space model.
\label{fig:knownkernel}}
\end{center}
\vskip -0.2in
\end{figure*}

To achieve an accurate approximation to a given target kernel, what rank of LEG model is necessary?  Unfortunately, we have been unable to obtain the convergence rates for Theorem \ref{lem:kdereg} that would enable a firm answer to this question; rates for total variation convergence of kernel density estimators can be challenging to obtain without additional assumptions.  To get some idea for what convergence rates can be expected, we turn to numerical experiments.

We considered several kernels: a squared exponential kernel, a triangle kernel, a rational quadratic kernel with $\alpha=2$, a sinc kernel, and a Matern kernel of order $1$.  Note that Matern kernels of half-integer order can be represented exactly with LEG kernels, but we do not have the same guarantee for integer-order Matern kernels such as the one considered here.  Indeed, as far as we are aware, none of these kernels can be represented exactly with any state-space model, but we will see that all of them can be approximated with reasonable accuracy using LEG kernels.

For each target kernel, we wanted to minimize the maximum absolute difference between the target kernel and a LEG kernels of various ranks.  However, this objective is difficult to optimize.  Fortunately, the total variation (``L1'') distance between the two kernels in Fourier space is an upper bound for the uniform error (indeed, if $M,\hat M$ are the spectra for $C,\hat C$, then observe that 
$
\sup_{\tau} |C_{ij}(\tau)-\hat C_{ij}(\tau)| \leq \sup_{\tau} \int |\exp(-i\omega\tau)| |M_{ij}(\omega) - \hat M_{ij}(\omega)| d\omega
$).  To obtain an approximating kernel, we thus started from a random initial condition and applied gradient descent on the total variation loss in the Fourier space.

In addition to this gradient-descent approach, we also used two other proposals from the literature.  \citet{sarkka2019applied} give a proposal based on Pad\'e approximants to approximate squared exponential kernels using state space models; per Lemma \ref{lem:smareleg}, this proposal can be interpreted as a particular choice of LEG kernel designed to approximate the squared exponential kernel.  Similarly, \citet{sarkka2019applied} give a proposal for a state-space representation of rational quadratic kernels using a mixture of approximate squared exponential kernels.

The results are summarized in Figure \ref{fig:knownkernel}.  With the exception of the triangle kernel, all kernels could be uniformly matched within $.01$ by a LEG kernel of rank 7; the triangle kernel could be uniformly matched within $.01$ by a LEG kernel of rank 13.   Given that the ground-truth kernel is rarely known exactly in any case, this suggests a rank-7 LEG kernel may often be sufficient in practice.  The triangle kernel was the most difficult to match, perhaps because of its discontinuous derivative.  In most cases the proposals from the literature yielded higher error than the LEG kernels found by gradient descent, though for rank-12 kernels gradient descent was unable to find a model with less error than the Pad\'e approximant.  
A cleverer optimization strategy might yield numerically stable low-rank LEG kernels that are even more accurate than the ones presented here.   In future we hope to investigate better strategies for initialization and optimization.

\subsection{Parameter estimation from simulated data}


Given a set of data drawn from a stationary Gaussian Process, Section \ref{sec:leggpcomputation} proposes to estimate the covariance of this process by using gradient descent to seek a maximum-likelihood LEG model for that data.  We will call this the ``maximum-likelihood-based LEG estimator,'' although the likelihood is not convex and gradient descent may not find the global maximum-likelihood estimator.   Will this estimator be accurate?  Theorem \ref{thm:legflex} suggests it might be, especially given unlimited data and sufficiently high choice of rank.  However, we have no theorem on this subject.  Any such theorem would require us to specify how the rank $\nlat$ should grow as the amount of data grows; in future we hope to explore this question more thoroughly.  For now, we turn to numerical experiments.  

We examine six kernels: the sum of a Radial Basis Function (RBF) and Rational Quadratic (RQ) kernels, an RBF kernel multiplied by cosine, the sum of an RBF kernel with one-fifth of an RQ kernel, the product of a cosine with the previous kernel, a triangle kernel, and the product of an RBF kernel with a high-frequency cosine.  In each case we draw eighteen-thousand observations, each taken $.1$ units apart from the next.  We then attempt to estimate the true covariance of the model from these observations.

\begin{figure*}
\vskip 0.2in
\begin{center}
\hfill\begin{minipage}{.9\columnwidth}
\includegraphics[width=\columnwidth]{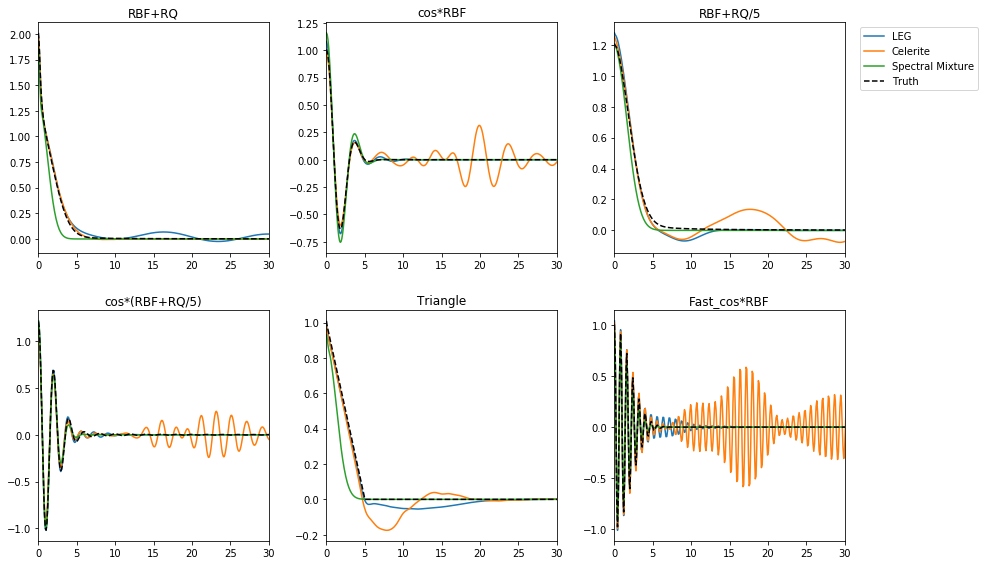}
\end{minipage}\hfill{}

\caption{\textbf{The maximum-likelihood-based LEG estimator is accurate in simulations.}  We tried to learn covariance kernels from data, using maximum likelihood with three different model families: LEG, Celerit\'e, and spectral mixtures of squared exponential kernels.  We found the LEG and spectral mixture families yielded comparable accuracy (though the spectral mixture methods are slower as they have no state-space representation); the Celerit\'e estimates performed a bit worse, exhibiting some spurious cyclic activity.  
\label{fig:paretos}}
\end{center}

\vskip -0.2in
\end{figure*}

We compare the maximum-likelihood-based LEG estimator with two other approaches: the Celerit\'e approach and an approach based on spectral mixtures of squared-exponential kernels.  We estimated Celerit\'e parameters using the \texttt{celerite2} package and estimated the SM parameters using the  \texttt{gpytorch} package.  All three families include a ``rank'' hyperparameter (i.e. the rank of LEG models, the number of terms in a Celerit\'e kernel, and the number of mixtures in an SM kernel); in order to give all models the best chance, we exhaustively searched over values of this rank hyperparameter and selected the model with the best likelihood on six thousand held-out samples (the best LEG kernels were of rank 7-11, the best Celerit\'e kernels were of rank 6-7, and the best SM kernels were of rank 4-8).  The Spectral Mixture kernels and the LEG kernels appear to perform comparably; the Celerit\'e family is designed for periodic astronomical phenomena, and has a corresponding inductive bias that leads it to estimate overlarge periodic components.  These results are summarized in Figure \ref{fig:paretos}.  The LEG and spectral mixture kernels perform comparably (though the spectral mixture kernels do not have a state-space representation and took longer to train), and the Celerit\'e approach (designed for periodic astronomical phenomena) exhibited some spurious cyclic activity.   In brief, we see no sign that the LEG family carries a problematic inductive bias for parameter estimation.

\subsection{Learning, smoothing, and interpolating on real-world data}

\subsubsection{Interpolation for the Mauna Loa CO$_2$ dataset}

\begin{figure*}[t!]
\vskip 0.2in
\begin{center}
\centerline{\includegraphics[width=\columnwidth]{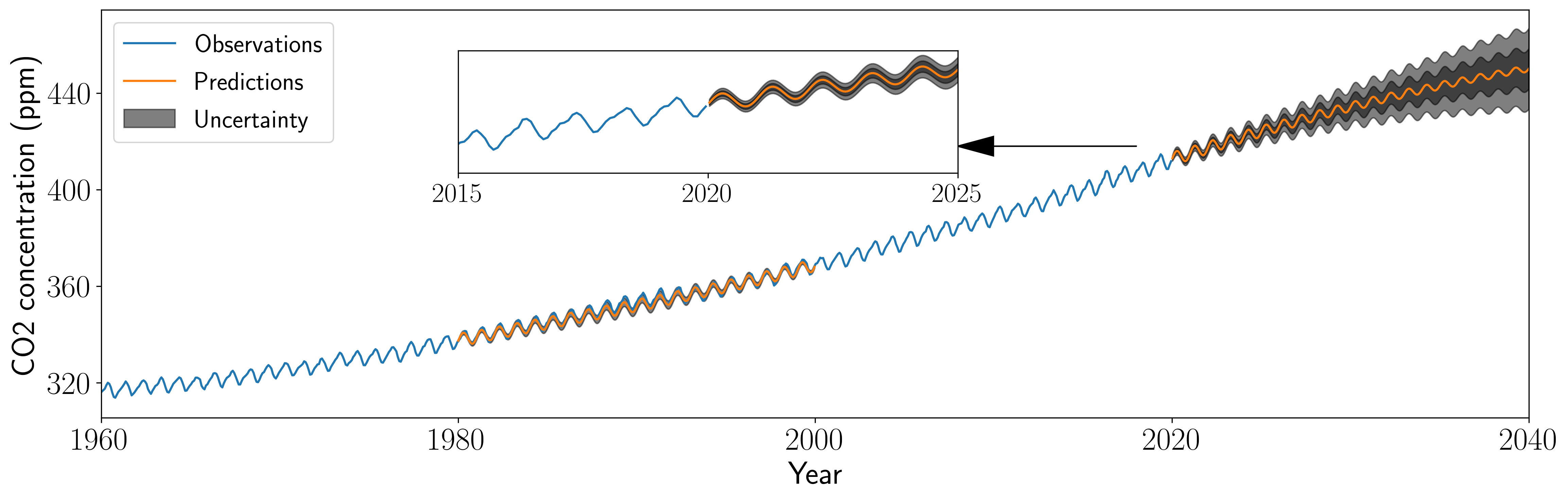}}
\caption{\textbf{LEG processes interpolate and extrapolate well across long timescales}.  It appears that a rank-5 LEG model is sufficient to capture the linear and periodic trends in the Mauna Loa CO$_2$ dataset.  Above we compare the true observations with interpolations made by the LEG model.  The gray areas encompass one and two predictive standard deviations, i.e. the LEG model's uncertainty in forecasting and extrapolating what out-of-sample observations would look like.}
\label{fig:realdata}
\end{center}
\vskip -0.2in
\end{figure*}

To check whether LEG parameterization can capture periodic covariances, we turned to the Mauna Loa CO$_2$ dataset.  For the last sixty years, the monthly average atmosphere CO$_2$ concentrations at the the Mauna Loa Observatory in Hawaii have been recorded \citep{keeling2005atmospheric}.  This dataset is interesting because it features two different kinds of structures: an overall upward trend and a yearly cycle.  To test the ability of the LEG model to learn these kinds of structures from data, we trained a rank-5 LEG kernel on all the data before 1980 and all the data after 2000.  We then asked the LEG model to interpolate what happened in the middle and forecast what the concentration might look like in the next twenty years.  

The results are shown in Figure \ref{fig:realdata}.  It is encouraging that the LEG predictions interpolate adequately from 1980 to 2000.  Even though the LEG process is given no exogenous information about  ``years'' or ``seasons,'' it correctly infers the number of number of bumps between 1980 and 2000.  This example shows that the LEG model is sufficiently flexible to learn unanticipated structures in the data.  

\subsubsection{Smoothing for a Hippocampal place-cell dataset}

\begin{figure*}
\vskip 0.2in
\begin{center}
\centerline{\includegraphics[width=\columnwidth]{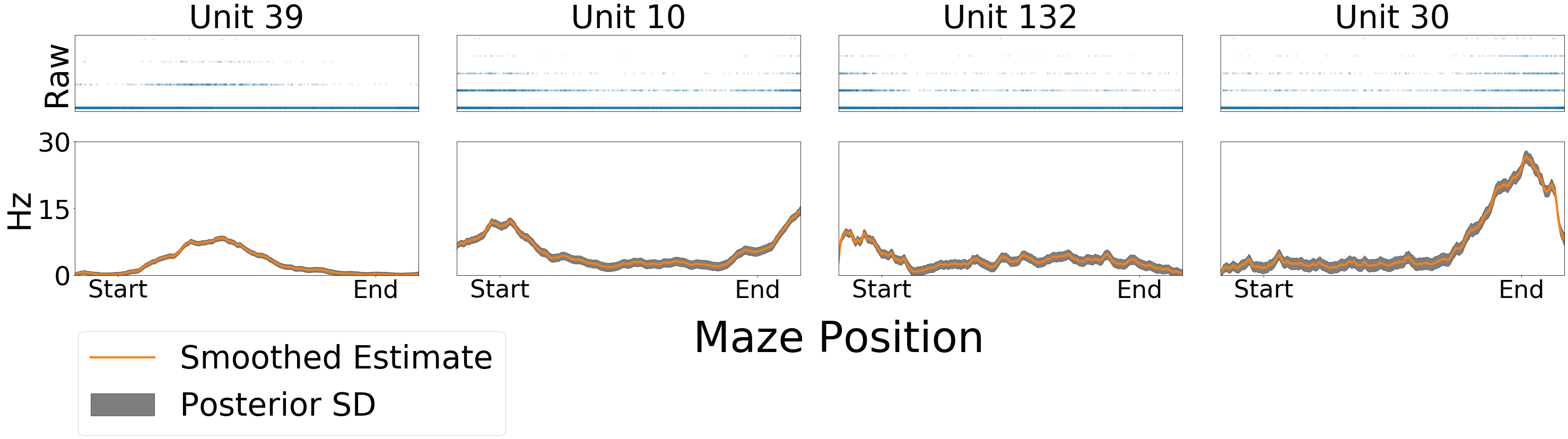}}
\caption{\textbf{LEG processes can smooth irregularly spaced neural data}.  Ten thousand irregularly spaced observations suggest that the firing rates of Hippocampal neurons are modulated by the rat's position in a maze.  However, the modulation strength visible in the raw data is weak.  By smoothing this data with a LEG process we can see the trend more clearly. How much should we smooth?  By training a rank-5 LEG process we can determine a smoothness level automatically. The gray areas indicate the LEG model's posterior uncertainty about the estimated tuning curve.}
\label{fig:kenny}
\end{center}
\vskip -0.2in
\end{figure*}

Finally, we wanted to check whether the LEG approach for covariance estimation might introduce any strange artifacts when applied to irregularly-spaced data.   

We looked at observations from neural spiking data \citep{grosmark2016diversity}.  In this data a rat's position in a one-dimensional maze is reported on a regular schedule, around 40 times per second.  At each time-step, each neuron may be silent or may fire (``spike'') some number of times.  For each neuron, we would like to estimate the ``tuning curve'' -- a function which takes in positions and returns the expected number of spikes as a function of the rat's position.  With no smoothness assumptions on this function, the problem is impossible; the rat is never observed at exactly the same place twice.  However, it is unclear how much smoothness should be assumed.  Gaussian Processes offer a natural way to automatically learn an appropriate level of smoothness from the data itself.  Note that the observed positions do not fall into a regularly spaced grid, so classical approaches such as the ARMA model cannot be directly applied.  

Here we model this tuning curve using a PEG process, $z\sim \PEGGP(N,R)$.  In this view, each data-point from the experiment constitutes a noisy observation of $\latV$.  When the rat is at position $t$  we model the distribution on the number of spikes observed in a small timebin as a Gaussian, with mean $B z(t)$ and variance $\Lambda \Lambda^\top$.  (It would be interesting to apply a non-Gaussian observation model here, as in, e.g., \citep{smith2003estimating,rad2010efficient,savin2016,gao2016linear}, and references therein; 
as noted in section \ref{sec:extensions}, linear-time approximate inference is feasible in this setting and is an important direction for future work.)  For each neuron we train the parameters of a separate LEG model.  We can then look at the posterior distribution on the underlying tuning curve $\latV$.  The posterior mean of this process for various neurons is shown in Figure \ref{fig:kenny}.  We also represent one standard-deviation of the posterior variance of $\latV$ with gray shading.   Fitting the LEG model and looking at the posterior under the learned model appears to yield an effective Empirical Bayes approach for this kind of data.

\subsection{Speed}

\begin{figure}[t!]
\vskip 0.2in
\begin{center}
\begin{minipage}{.9\columnwidth}
\includegraphics[width=\columnwidth]{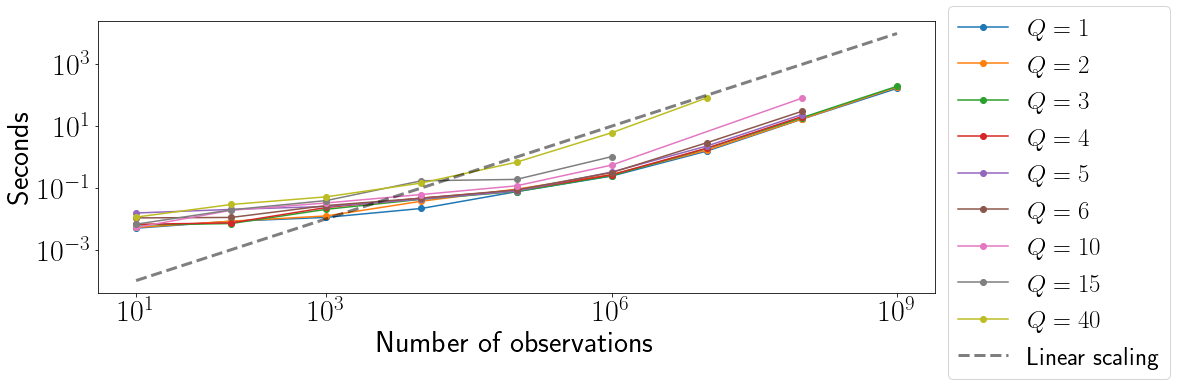}
\end{minipage}

\begin{minipage}{1.0\columnwidth}
\caption{\textbf{Walltime for evaluating PEG likelihoods}.  How long does it take to compute the likelihood for a PEG model on an m5.24xlarge machine on the Amazon AWS service?   We compare times for different ranks ($\nlat$) and different numbers of blocks (\# Observations).   For example, the likelihood for one billion observations with $\nlat=3$ can be computed in three and a half minutes.  
\label{fig:sofast}}
\end{minipage}
\end{center}
\vskip -0.2in
\end{figure}

In theory, the algorithms proposed in Section \ref{sec:leggpcomputation} should scale linearly with the number of samples.  However, we wanted to check if there might be constant computational costs that might make this result irrelevant in practice.  To do so, we measured how long it took to compute the likelihood for a PEG model.  
We varied both the number of observations and the rank.   In each case we used an m5-24xlarge machine on Amazon Web Services (AWS).  

Overall, the empirical scaling appeared consistent with the theoretical predictions of linear scaling.  The likelihood of one million observations from a rank-3 model could be computed in $0.25$ seconds, and one billion observations could be computed in 195 seconds.  We saw similar trends across models of other ranks; the results are summarized in Figure \ref{fig:sofast}.  Note that for smaller datasets we actually observed a sublinear scaling (i.e. a slope of less than one on the log-log plot) that turns linear for larger values of $m$.

We also wanted to check whether the algorithms proposed in Section \ref{sec:leggpcomputation} would make adequate use of parallel hardware.  We compared the \texttt{leggps} package with two alternatives: \texttt{pylds} (a single-threaded C package for state-space models) and \texttt{pyro.distributions.GaussianHMM} (a parallel associative-scan package for state-space models).  We also investigated how performance varied between CPU-only hardware and GPU hardware.  These experiments yielded three main findings:

\begin{itemize}
    \item GPU hardware is faster than any number of CPU cores -- but with enough samples it becomes impossible to keep everything in GPU memory, so it is not easy to use GPUs to directly compute likelihoods on billions of observations.  If the process is expected to mix well within a few million observations, this difficulty can be circumvented by using batches of a few million observations at a time; it may even be possible to learn the parameters using a subset of the data.  Once the model is learned on the GPU using this batch-approach, exact smoothing for a much larger dataset can be performed on the CPU.   However, if the process has extremely long-range dependencies, new approaches would be necessary. One option would be to break the data into tiles and move the relevant tensors in and out of GPU memory as necessary to compute the relevant quantities.  We hope to explore these possibilities in the future.  
    \item CR and associative-scan methods seem to have roughly the same runtimes.  Likely as not, any difference in runtime reflects a difference in implementation details (e.g. the \texttt{GHMM} package was written in PyTorch, but \texttt{leggps} was written in TensorFlow).
    \item When the number of observations is less than 200, the single-threaded \texttt{pylds} package can be slightly faster.  
\end{itemize}

To obtain these results, we used two different measures of speed.  To compare CR with associative-scans, we evaluated the time required to compute the gradient of the likelihood of a state-space model with respect to its parameters.  To compare with \texttt{pylds}, we evaluated the time required to compute posterior variances.  We did not use the same benchmark for both packages because the \texttt{pylds} package does not compute gradients and the \texttt{GaussianHMM} package does not compute posterior variances.  

\begin{figure*}[t!]
\begin{center}
\centerline{\includegraphics[width=\columnwidth]{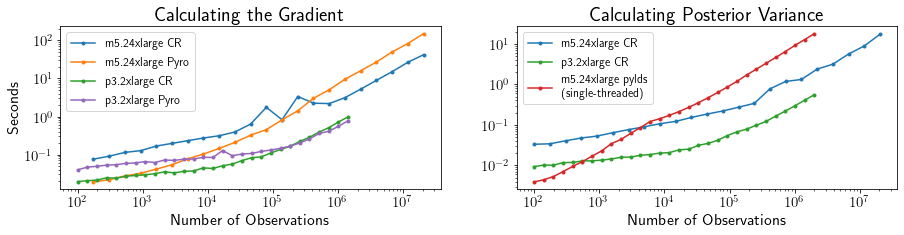}}
\caption{\textbf{\texttt{leggps} computes with state-of-the-art speed}.  
Here we evaluate time required to perform two key tasks for inference and learning with state-space models.  In each case we assume $\nlat=5,\ndim=4$; we found similar results for other dimensionalities.  We look at three different packages: \texttt{pyro.distributions.GaussianHMM} (Pyro), \texttt{pylds} (pylds), and \texttt{leggps} (CR).  We also look at two different types of hardware on Amazon ec2:  m5.24xlarge machines (96 cores with 384 gigabytes of RAM) and p3.2xlarge (one Nvidia V100 GPU).  These plots suggest three overall findings: Pyro and CR perform comparably, a single GPU is much faster than any number of CPUs, and pylds can be faster when the number of observations is small.
}
\label{fig:practicalcomp}
\end{center}
\end{figure*}

The times are summarized in Figure \ref{fig:practicalcomp}.  Several remarks are in order about these plots:

\begin{itemize}
    \item When the number of observations is large, the GPU architecture is unable to finish the computations due to RAM limitations.
    \item We did not run pylds on very large datasets due to the prohibitive time costs.
    \item We used Amazon hardware to perform our tests.  We used m5.24xlarge machines to represent CPU computation (96 cores from Intel Xeon Platinum 8000 processors, 384 gigabytes of memory), and p3.2xlarge machines to represent GPU computation (one Nvidia Tesla V100 processor with 16 gigabytes of memory).  Note that the \texttt{pylds} package is single-threaded in nature, and unable to take advantage of the 96 cores offered by the m5.24xlarge; this package achieves essentially the same times even with much more modest machines such as the m5.large machine.   
\end{itemize}

\section{Conclusions}

We here provide several advances for Gaussian Processes on one dimension.  First, we show that the LEG model (a particularly tractable continuous-time Gaussian hidden Markov process) can be used to approximate any GP with a stationary integrable continuous kernel, enabling linear runtime scaling in the number of observations.  Second, we develop a new unconstrained parameterization of continuous-time state-space models, making it easy to use simple gradient-descent methods.  Finally, we develop a simple package for learning, interpolation, and smoothing with such models; this package uses Cyclic Reduction to achieve rapid runtimes on modern hardware.  We believe these advances will open up a wide variety of new applications for GP modeling in highly data-intensive areas involving data sampled at high rates and/or over long intervals, including geophysics, astronomy, high-frequency trading, molecular biology, neuroscience, and more.

\section*{Acknowledgements}

Thanks to Jake Soloff for resolving a thorny point about matrix-valued measures.

We gratefully acknowledge support from NIH U19NS107613 (JL+LP), IARPA D16PC00003 (JL+LP), NSF NeuroNex (JL+JC+LP), the Simons Foundation (JC+LP) and the Gatsby Charitable Foundation (JL+JC+LP). David Blei was supported by ONR N00014-17-1-2131, ONR N00014-15-1-2209, NSF CCF-1740833, DARPA SD2 FA8750-18-C-0130, 2Sigma, Amazon, and NVIDIA. John P.\ Cunningham was additionally supported by the McKnight Foundation.

\bibliographystyle{plainnat} 
\bibliography{refs}

\appendix


\section{Computations with block-tridiagonal matrices using Cyclic Reduction
(CR)}

Let $J$ be the symmetric positive-definite block-tridiagonal matrix,
defined blockwise by 
\[
J=\left(\begin{array}{cccc}
R_{0} & O_{0}^{T} & 0 & \cdots\\
O_{0} & R_{1} & O_{1}^{T}\\
0 & O_{1} & R_{2}\\
\vdots &  &  & \ddots
\end{array}\right)
\]
We would like to be able to compute efficiently with $J$. To do so,
we can decompose $J$ using what is called a ``Cyclic Reduction'' ordering.  This ordering is well-known among those well-acquainted with parallel matrix algorithms \cite{sweet1974generalized}.  However, as far as we are aware the literature currently lacks a precise specification of how this ordering can be used for the particular algorithms needed in our application.  For the benefit of the reader, we present a self-contained summary here.

\begin{definition}[Cyclic Reduction]
For each $m$, let $P_{m}$
\[
P_{m}=\left(\begin{array}{cccccc}
I & 0 & 0 & 0 & 0\\
0 & 0 & I & 0 & 0\\
0 & 0 & 0 & 0 & I\\
 &  &  &  &  & \ddots
\end{array}\right)
\]
denote the permutation matrix which selects every other block of a
matrix with $m$ blocks. Let 
\[
Q_{m}=\left(\begin{array}{cccccc}
0 & I & 0 & 0 & 0\\
0 & 0 & 0 & I & 0\\
0 & 0 & 0 & 0 & 0\\
 &  &  &  &  & \ddots
\end{array}\right)
\]
denote the complementary matrix, which takes the other half of the
blocks.

The \textbf{Cyclic Reduction} of a block-tridiagonal matrix $J$ with $m$
blocks is defined recursively by 
\begin{align*}
L=\CyclicReduction\left(J,m\right) & =\left(\begin{array}{cc}
P_{m}^{T} & Q_{m}^{T}\end{array}\right)\left(\begin{array}{cc}
D & 0\\
U & \tilde L
\end{array}\right)\\
D & =\chol\left(P_{m}JP_{m}^{\top}\right)\\
U & =Q_{m}JP_{m}^{T}D^{-\top}\\
\tilde{J} & =Q_{m}JQ_{m}^{T}-U^{\top}U\\
\tilde{L} &= \CyclicReduction\left(\tilde{J},\left\lceil m/2\right\rceil \right)
\end{align*}
Note that the $P,Q$ matrices select subsets of blocks of the matrices involved.  For example, it is straightforward to show that $P_m J P_m^\top$ is block-diagonal, with blocks given by $R_0,R_2,R_4,\cdots$.  The matrix $Q_m J Q_m^\top$ is also block-tridiagonal, with blocks given by $R_1,R_3,R_5,\cdots$.  The matrix $Q_m J P_m^\top$ is upper block-didiagonal (i.e. it has diagonal blocks and one set of upper off-diagonal blocks); the diagonal blocks are given by $O_0,O_2,O_4,\cdots$ and the upper off-diagonal blocks are given by $O_1,O_3,O_5,\cdots$.

For this recursive algorithm to make sense, we need that $\tilde J$ is also block-tridiagonal -- but this is always true if $J$ is block-tridiagonal.  The recursion terminates when $J$ has exactly one block.  For this we define the base-case
\[
\CyclicReduction\left(J,1\right) = \chol(J)
\]
\end{definition}

\begin{prop}
Let $L=\CyclicReduction\left(J,n\right)$.  Then $LL^\top = J$.
\end{prop}
\begin{proof}
By induction.  For the case $n=1$ the algorithm works because the Cholesky decomposition works.  

Now let us assume the algorithm works for all $\tilde n<n$.  We will show it works for $m$.  Let
\begin{align*}
L&=\left(\begin{array}{cc}
P_{m}^{T} & Q_{m}^{T}\end{array}\right)\left(\begin{array}{cc}
D & 0\\
U & \tilde L
\end{array}\right)\\
D & =\chol\left(P_{m}JP_{m}^{\top}\right)\\
U & =Q_{m}JP_{m}^{T}D^{-\top}\\
\tilde{J} & =Q_{m}JQ_{m}^{T}-U^{\top}U\\
\tilde{L} &= \CyclicReduction\left(\tilde{J},\left\lceil m/2\right\rceil \right)
\end{align*}
By induction $\tilde L \tilde L^\top = \tilde J$.  Thus
\begin{align*}
LL^{T}&=\left(\begin{array}{cc}
P_{m}^{T} & Q_{m}^{T}\end{array}\right)\left(\begin{array}{cc}
D & 0\\
U & \tilde{L}
\end{array}\right)\left(\begin{array}{cc}
D^{T} & U^{T}\\
0 & \tilde{L}^{T}
\end{array}\right)\left(\begin{array}{c}
P_{m}\\
Q_{m}
\end{array}\right)\\&=\left(\begin{array}{cc}
P_{m}^{T} & Q_{m}^{T}\end{array}\right)\left(\begin{array}{cc}
DD^{T} & DU^{T}\\
UD^{T} & UU^{T}+\tilde{L}\tilde{L}^{T}
\end{array}\right)\left(\begin{array}{c}
P_{m}\\
Q_{m}
\end{array}\right)\\&=\left(\begin{array}{cc}
P_{m}^{T} & Q_{m}^{T}\end{array}\right)\left(\begin{array}{cc}
P_{m}JP_{m}^{T} & \cancel{DD^{-1}}P_{m}JQ_{m}^{T}\\
Q_{m}JP_{m}^{T}\cancel{D^{-T}D^{T}} & \cancel{UU^{T}}+Q_{m}JQ_{m}^{T}-\cancel{UU^{T}}
\end{array}\right)\left(\begin{array}{c}
P_{m}\\
Q_{m}
\end{array}\right)\\&=J
\end{align*}
\end{proof}

This decomposition enables efficient computations with $J$.  Below we describe all of the relevant algorithms (including the CR decomposition algorithm itself) from an algorithms point of view, giving runtimes for each.  We will see that all operation counts scale linearly in the number of blocks.  We will also discuss parallelization; as we shall see, almost all of the work of a Cyclic Reduction iteration can be done in parallel across the $m$ blocks of $J$.   

\subsection{Cyclic Reduction}

\begin{algorithm}[H]
\SetKwInOut{Input}{input}
\SetKwInOut{Output}{output}
 \Input{rblocks,oblocks,$m$ -- the diagonal and lower off-diagonal blocks of a block-tridiagonal matrix $J$ which has $m$ blocks}
 \Output{dlist,flist,glist-- a representation of the CR decomposition of $J$}
 \eIf{$m=1$}{
    \Return $[\chol(R_0)],[],[]$
 }
 {
    Adopt the notation $R_i=$rblocks[i] and $O_i=$oblocks[i]\;
    Let
    \[
    D \triangleq  \left(\begin{array}{ccc}
    D_0 & 0 & 0\\
    0 & D_1 \\
     &  & \ddots
    \end{array}\right) 
    \triangleq \left(\begin{array}{ccc}
    \chol\left(R_{0}\right) & 0 & 0\\
    0 & \chol\left(R_{2}\right)\\
     &  & \ddots
    \end{array}\right)
    \]
    and store the diagonal blocks of $D$ in dblocks\;
    Let
    \[
    U \gets \left(\begin{array}{ccccc}
    O_{0}D_{0}^{-T} & O_{1}D_{1}^{-T} & 0 & \cdots & 0\\
    0 & O_{2}D_{1}^{-T} & O_{3}D_{2}^{-T}\\
    0 & 0 & O_{4}D_{2}^{-T} & \ddots\\
    \vdots &  &  & \ddots & \\
    \end{array}\right)
    \]
    and store diagonal and upper-off-diagonal blocks of $U$ in (fblocks,gblocks)\;
    Let
    \[
    \tilde J = \left(\begin{array}{ccc}
    R_1 & 0 & 0\\
    0 & R_3\\
     &  & \ddots
    \end{array}\right) - UU^\top
    \]
    and store the diagonal and lower-offdiagonal blocks of $\tilde J$ in newrblocks,newoblocks\;
    newdlist,newflist,newglist $\gets$ decompose(newrblocks,newoblocks,len(newrblocks))\; 
    \Return concat([dblocks],newdlist),concat([fblocks],newflist),concat([gblocks],newglist)\;
 }
 \caption{decompose}
\end{algorithm}

Observe that the dlist,flist,glist returned by this algorithm stores everything we would need to reconstruct the $\CyclicReduction(J)$.

How long does this algorithm take?
\begin{itemize}
    \item Step 5 requires we compute $m$ Cholesky decompositions
    \item Step 6 requires $m-1$ triangular solves
    \item Step 7 has two components.  First we must compute the diagonal and lower-off-diagonal blocks of $UU^\top$ (which requires about $m$ matrix-multiplies and $m$ matrix additions).  Second we must compute $\lfloor m/2 \rfloor$ matrix subtractions.
    \item Step 8 requires we run the CR algorithm on a problem with $\lfloor m/2 \rfloor$ blocks.
\end{itemize}

Let $C(m)$ denote overall number of operations for a Cyclic Reduction on an $m$-block matrix.  Since steps 7, 8, and 9 require $O(m)$ operations, we have that there exists some $c$ such that
\[
C(m) \leq cm + C(\lfloor n/2 \rfloor) \qquad C(1) \leq c
\]
from which we see that $C(m) < 2cm$,\footnote{One way to see this is by induction.  For the base case, we have $C(1)\leq c$.  Then, under the inductive hypothesis, we have that $C(m) < c(m+2m/2) = 2cm$.  In general for all the recursive algorithms that follow, to prove linear-time it will suffice to show that the non-recursive steps require $O(m)$ time.} i.e. the computation scales linearly in $m$.

What about parallelization?  To compute steps 5-7 we need to compute many small Cholesky decompositions, compute many small triangular solves, compute many small matrix multiplies.  These are all common problems, and fast algorithms exist for achieving these goals on multiple CPU cores or using GPUs.

\subsection{Solving $Lx=b$}

This algorithm uses the tuple (dlist,flist,glist) representing a Cyclic Reduction $L$ on a matrix with $m$ blocks to compute $L^{-1}b$.

\begin{algorithm}[H]
\SetKwInOut{Input}{input}
\SetKwInOut{Output}{output}
 \Input{dlist,flist,glist,$b$,$m$}
 \Output{$x=L^{-1}b$}
 Adopt the notation $D$ is the block-diagonal matrix whose diagonal blocks are given by dlist[0] \;
 Adopt the notation that $U$ is the upper didiagonal matrix whose diagonal blocks are given by flist[0] and whose upper off-diagonal blocks are given by glist[0]\;
 \eIf{$m=1$}{
     \Return $x=D^{-1} b$
 }
 {
    $x_1 \gets D^{-1} P_m b$\;
    $x_2 \gets$ halfsolve(dlist[1:],flist[1:],glist[1:],$Q_m b - U x_1$,$\lfloor m/2\rfloor$)\;
    \Return 
    \[
        x=\left(\begin{array}{c}
        x_1\\
        x_2
        \end{array}\right)
    \]
 }
 \caption{halfsolve}
\end{algorithm}

Note that step 6 requires $O(m)$ operations, the base case requires $O(1)$ operations, and step 7 is a recursion on a problem of half-size.  The overall computation thus scales linearly in $m$.  Moreover, step 6 can be understood as $m$ independent triangular solves, all of which can be solved completely independently (this algorithm is thus easy to parallelize across many cores).  

\subsection{Solving $L^{\top}x=b$}

This algorithm uses the tuple (dlist,flist,glist) representing a Cyclic Reduction $L$  on a matrix with $m$ blocks to compute $L^{-\top}b$.

\begin{algorithm}[H]
\SetKwInOut{Input}{input}
\SetKwInOut{Output}{output}
 \Input{dlist,flist,glist,$b$,$m$}
 \Output{$x=L^{-\top}b$}
 Adopt the notation $D$ is the block-diagonal matrix whose diagonal blocks are given by dlist[-1] (i.e. the last entry in dlist)\;
 Adopt the notation that $U$ is the upper didiagonal matrix whose diagonal blocks are given by flist[-1] and whose upper off-diagonal blocks are given by glist[-1]\;
 \eIf{$m=1$}{
     \Return $x=D^{-T} b$
 }
 {
    $\tilde x_2 \gets$ backhalfsolve(dlist[:-1],flist[:-1],glist[:-1],$b$,$\lfloor m/2\rfloor$)\;
    $\tilde x_1 \gets D^{-\top} (P_n b- U^\top \tilde x_2)$\;
    \Return 
    \[
    x=\left(\begin{array}{cc}
    P_{n}^{\top} & Q_{n}^{\top}\end{array}\right)\left(\begin{array}{c}
    \tilde{x}_{1}\\
    \tilde{x}_{2}
    \end{array}\right)
    \]
 }
 \caption{backhalfsolve}
\end{algorithm}

Just like the halfsolve algorithm, the cost of backhalfsolve scales linearly in $m$ and is easy to parallelize across the $m$ blocks.

\subsection{Solving $Jx=b$}

\begin{enumerate}
\item First solve $Ly=b$ using halfsolve.
\item Then solve $L^{T}x=y$ using backhalfsolve.
\end{enumerate}
Then 
\[
Jx=LL^{T}x=Ly=b
\]
as desired.

\subsection{Computing determinants}

The determinant of a block-Cholesky decomposition is just the square of the product of the determinants of the diagonal blocks.  Thus if (dlist,flist,glist) represents the CR decomposition of $J$ we have that the determinant of $J$ is given by the square of the products of the determinants of all the matrices in dlist.  This can be done in parallel across all of the $m$ blocks, requiring $O(m)$ operations in total.

\subsection{Computing the diagonal and off-diagonal blocks of the inverse}

\begin{algorithm}[H]
\SetKwInOut{Input}{input}
\SetKwInOut{Output}{output}
 \Input{dlist,flist,glist}
 \Output{diags,offdiags -- the diagonal and off-diagonal blocks of $J$}
 Adopt the notation $D$ is the block-diagonal matrix whose diagonal blocks are given by dlist[-1] (i.e. the last entry in dlist)\;
 Adopt the notation that $U$ is the upper didiagonal matrix whose diagonal blocks are given by flist[-1] and whose upper off-diagonal blocks are given by glist[-1]\;
 \eIf{$m=1$}{
     \Return [$D^{-T} D^{-1}$],[]
 }
 {
    subd,suboff $\gets$ invblocks(dlist[:-1],flist[:-1],glist[:-1])\;
    Adopt the notation that $\tilde \Sigma$ is a matrix whose diagonal blocks are given by subd and whose lower off-diagonal blocks are given by suboff\;
    Let SUDid store the diagonal blocks of $\tilde\Sigma U D^{-1}$\;
    Let DitUtSo store the upper-off-diagonal blocks of $D^{-\top}U^{\top}\tilde\Sigma$\;
    Let DitUtSUDid store the diagonal blocks of $D^{-\top} U^\top \tilde\Sigma U D^{-1}$\;
    Let 
    \[
    \mathrm{diags} \gets \left(\begin{array}{cc}
    P_{n}^{\top} & Q_{n}^{\top}\end{array}\right)\left(\begin{array}{c}
    \mathrm{DitUtSUDid}\\
    \mathrm{subd}
    \end{array}\right)
    \]
    where DitUtSUid and subd are understood as tall columns of matrices.  For example, if each block is $\ell \times \ell$, we understand DitUtSUDid as a $\lceil m/2\rceil \times \ell$ matrix\;
    Let
    \[
    \mathrm{offdiags} \gets
    \left(\begin{array}{cc}
    P_{n}^{\top} & Q_{n}^{\top}\end{array}\right)\left(\begin{array}{c}
    \mathrm{SUDid}\\
    \mathrm{DitUtSo}
    \end{array}\right)
    \]
    where SUDid and DitUtSo are understood as tall columns of matrices\;
    \Return diags,offdiags\;
 }
 \caption{invblocks}
\end{algorithm}

The cost of this algorithm scales linearly in $m$ because steps 7-11 require $O(m)$ operations.  To see how this can be so, recall that $U$ is block didiagonal and $D$ is block diagonal.  Thus, for example, step 8 involves $\Sigma U D^{-1}$.   Computing this entire matrix would be quite expensive.  However, $U$ is block didiagonal and we only need the diagonal blocks of the result, so we can get what we need in linear time and we only need to know the diagonal and off-diagonal blocks of $\Sigma$.  As in the other algorithms, note that all of the steps can be done in parallel across the $m$ blocks.

Why does this algorithm work?  As usual, let
\begin{align*}
L=\CyclicReduction\left(J,m\right) & =\left(\begin{array}{cc}
P_{m}^{T} & Q_{m}^{T}\end{array}\right)\left(\begin{array}{cc}
D & 0\\
U & \tilde L
\end{array}\right)\\
D & =\chol\left(P_{m}JP_{m}^{\top}\right)\\
U & =Q_{m}JP_{m}^{\top}D^{-\top}\\
\tilde{J} & =Q_{m}JQ_{m}^{\top}-U^{\top}U\\
\tilde{L} &= \CyclicReduction\left(\tilde{J},\left\lceil m/2\right\rceil \right)
\end{align*}
Now let $\tilde \Sigma = \tilde J^{-1}$.  It follows that 
\[
J^{-1} = \left(\begin{array}{c}
P_{n}\\
Q_{n}
\end{array}\right)\left(\begin{array}{cc}
D^{-T}D^{-1}+D^{-\top}U^{\top}\tilde{\Sigma}UD^{-1} & -D^{-\top}U^{\top}\tilde{\Sigma}\\
-\tilde{\Sigma}UD^{-1} & \tilde{\Sigma}
\end{array}\right)\left(\begin{array}{cc}
P_{n}^{T} & Q_{n}^{\top}\end{array}\right)
\]
So to compute the diagonal and off-diagonal blocks of $J^{-1}$ we just need to collect all the relevant blocks from the the inner matrix on the RHS of the equation above.  However, all of the relevant blocks can be calculated using only the diagonal and off-diagonal blocks of $\tilde \Sigma$.  This is what the invblocks algorithm does.

\subsection{Gradients of matrix exponentials}

\label{sec:apdxgrad}

In order to learn LEG models via gradient descent we will need to be able to compute gradients with respect to the parameters.  Most of these gradients can be computed automatically through a package such as TensorFlow.  There is one gradient which standard packages will compute inefficiently.  For many different values of $t$, we need to compute the value of $\exp(-tG)$.  By default, most packages will simply compute all the matrix exponentials for each $t$ separately, which is quite expensive.  However, all these matrix exponentials can be computed much more efficiently if $G$ is first diagonalized.  On the other hand, differentiating through diagonalization isn't very stable.  To get the gradients we need in a fast and stable manner we need to special-case the gradient-computation algorithm using Theorem 4.8 of  \citet{najfeld1995derivatives}.  We outline the relevant algorithm, below.

\begin{algorithm}[H]
\SetKwInOut{Input}{input}
\SetKwInOut{Output}{output}
 \Input{$G$ (a square matrix), tlist (a list of scalars)}
 \Output{rez (the value of $\exp(tG)$ for each $t$ in tlist), grad (a corresponding gradient-evaluating function)}
 $\Lambda,U\gets$ eigendecomposition$(G)$\;
 $\mathrm{Ui} \gets U^{-1}$\;
 $\mathrm{rez_{mjk}} \gets \sum_r U_{jr} \mathrm{Ui}_rk \exp(\Lambda_r \mathrm{tlist}_m)$\;
 grad $\gets \left(M \mapsto \mathrm{expmgrad}(\Lambda,U,\mathrm{Ui},\mathrm{rez},M)\right)$
 \caption{expm}
\end{algorithm}

\begin{algorithm}[H]
\SetKwInOut{Input}{input}
\SetKwInOut{Output}{output}
 \Input{$G,\Lambda$,$U$,Ui,rez,$M$,tlist}
 \Output{gradG ($mjk \mapsto \sum_{jk} M_{kr}(\partial \exp(\mathrm{tlist_m} G)_{kr}/\partial G_{ij}$)} and gradt ($m \mapsto \sum_{jk} M_{kr}(\partial \exp(t G)_{kr}/\partial t|_{t=\mathrm{tlist_m}}$)
 
 $\Phi_{mij} \gets \lim_{\lambda \rightarrow \Lambda_{i}}(\exp(\lambda \mathrm{tlist_m})-\exp(\Lambda_{j} \mathrm{tlist_m})/(\lambda - \Lambda_j)$\;
 $H_{ij} \gets \sum_m \Phi_{mij} (U^\top M_{m} \mathrm{Ui}^\top)_{ij}$\;
 $\mathrm{sand} \gets \mathrm{Ui}^\top H U^\top$\;
 \Return $\Re (\mathrm{sand})$\;
 \caption{expmgrad}
\end{algorithm}


\section{{Matrix-multiplication by Kronecker-Hadamard product of PEG kernels on a grid}}

\label{sec:mmkronhad}

The ideas introduced in this paper may also facilitate inference for GPs of the form $\mathbb{R}^d \rightarrow \mathbb{R}^D$.  For example, efficient computation is possible for kernels of the separable form $\tau \mapsto \prod_{k=1}^d C_\LEG(\tau_k;N_{k},R_{k},B_k)$; efficient matrix multiplication by the covariance matrices induced by sums of such kernels can be achieved by taking advantage of the special structure of the constituent LEG kernels, as shown in the following theorem.  

\begin{theorem} 
Let $\Omega = \prod_k^d \{\tau_{k1},\tau_{k2} \cdots \tau_{km} \} \subset \mathbb{R}^d$ denote a grid and $\{N_1\cdots N_d\},\{R_1\cdots R_d\},\{B_1\cdots B_d\}$ denote collections of $\nlat\times \nlat$ matrices.  For each $k$ let $K_k$ denote the $m \times \nlat \times m \times \nlat$ tensor such that $K_k(u,i,v,j)$ is equal to the value at row $i$ and column $j$ of $C_{\mathrm{LEG}}(\tau_{ku}-\tau_{kv};N_k,R_k,B_k)$.  Let $\Sigma$ denote the $m \times m \cdots \times \nlat \times m \times m \cdots \times \nlat$ tensor such that
\[
\Sigma(u_1,u_2,\cdots u_d,i,v_1,v_2 \cdots v_d,j) = \prod_{k=1}^d K_k(u_k,i,v_k,j).
\]
Let $x$ denote any $m \times m \cdots \times \nlat$ tensor.  Let $y$ denote the $m \times m \cdots \times \nlat$ tensor defined by
\[
y(u,i) = \sum_{v,j} \Sigma(u,i,v,j) x(v,j).
\]
For any fixed $N,R,B$, the cost of evaluating $y$ is $O(m^d)$ as a function of the width $m$ of the grid $\Omega$.
\end{theorem}
\begin{proof}
We compute $y$ by applying the operators $K_1, \cdots, K_d$ one at a time. 
\begin{align*}
y(u,i) 
 &= \sum_j^\nlat \sum_{v_1}^m\cdots \sum_{v_d}^m \left(\prod_k K_k(u_k,i,v_k,j)\right) x(v,j)\\
 &= \sum_j^\nlat \sum_{v_1}^m\cdots \sum_{v_{d-1}}^m 
   \left(\prod_k^{d-1} K_d(u_k,i,v_k,j)\right)
   \underbrace{\sum_{v_d}^m K_k(u_d,i,v_d,j) x(v,j)}_{\triangleq \xi_d(v_1\cdots v_{d-1},u_d,i,j) }
\end{align*}
If $K_d$ had no special structure, computing the $m \times m \times \cdots \nlat \times \nlat$ tensor $\xi$ would require $m^{d-1}$ matrix-vector multiplications, each costing $O(m^2)$.  However, as shown above, multiplications with with LEG kernels can be achieved with cost $O(m)$.   It follows that $\xi$ can in fact be computed with $O(m^d)$ cost.  Once this is computed, we proceed as follows.
\begin{align*}
y(u,i) 
 &= \sum_j^\nlat \sum_{v_1}^m\cdots \sum_{v_{d-2}}^m 
   \left(\prod_k^{d-2} K_k(u_k,i,v_k,j)\right) \underbrace{\sum_{v_{d-1}}^m K_{d-1}(u_{d-1},i,v_{d-1},j)\xi_d(v_1\cdots v_{d-1},u_d,i,j)}_{\triangleq \xi_{d-1}(v_1,v_2 \cdots u_{d-1},u_d,i,j})\\
\end{align*}
As before, computing the tensor $\xi_{d-1}$ requires only a $O(m^d)$ cost.  

We can continue in this way for $d$ steps, finally arriving at $y$ with $O(m^d)$ cost, as desired.  
\end{proof}

Therefore, matrix-vector multiplication with these separable multi-dimensional LEG kernels scales linearly with the number of grid points; it thus follows trivially that matrix-vector multiplication with finite sums of these separable multi-dimensional LEG kernels will also scale linearly with the number of grid points.

\section{Known results}
\label{sec:ap_known}

\subsection{Spectra of Gaussian Processes of the form $\obsV: \reals \rightarrow \reals^\nobs$}

Let $\Sigma:\ \reals^d \rightarrow \reals^{\nobs \times \nobs}$ denote the covariance kernel of some Gaussian Process.  The spectrum of $\Sigma$ admits a few properties which we use in our results.  

\begin{lemma}[Spectra of Gaussian Processes]\hspace{1in}

\begin{itemize}
    \item There exists a complex matrix-valued measure $\tilde M(d\omega)$ such that 
    \[
    \Sigma(\tau) = \int \exp(-i\tau \cdot \omega) \tilde M(d\omega)
    \]
    \item For any interval $(a,b)$, $\tilde M(a,b)$ is Hermitian and  nonnegative-definite.
    \item If $\Sigma$ is continuous and Lebesgue-integrable, $\tilde M$ is absolutely continuous with respect to Lebesgue measure on $\omega$.  In particular, it admits the representation $\tilde M(d\omega) = f(\omega)M(\omega)d\omega$ for a bounded probability density $f$ and a bounded Hermitian nonnegative-definite-valued function $M$.
\end{itemize}\end{lemma}
\begin{proof}
See \cite{cramer1940theory}.
\end{proof}
\subsection{Covariance and Spectra of Ornstein-Uhlenbeck Processes}

\begin{lemma}
Let $\latV$ denote a stationary zero-mean process satisfying
$
\latV(t) = \latV(0) + \int_0^t\left(-\frac{1}{2}G \latV(s)ds + N d\brownianmotion(s)\right),
$
where $\brownianmotion$ is a standard Brownian motion.  Then 
\begin{enumerate}
    \item The stationary covariance,  $P_\infty=\mathbb{E}[\latV(0)\latV(0)^\top]$, satisfies
    \[
    \frac{1}{2}\left(GP_{\infty}+P_{\infty}G^{\top}\right)=NN^{\top}.
    \]
    \item The covariance $C(\tau)=\mathbb{E}[\latV(\tau)\latV(0)^\top]$ is given below.
    \[
    C(\tau)=\begin{cases}
    \exp\left(-\frac{1}{2}G\left|\tau\right|\right)P_{\infty} & \text{if }\tau\geq0\\
    P_{\infty}\exp\left(-\frac{1}{2}G^{\top}\left|\tau\right|\right) & \text{if }\tau\leq0
    \end{cases}
    \]
    \item Define the spectrum $S$ as the unique function satisfying $C(\tau)=\int_{-\infty}^{\infty}S(\omega)\exp\left(-i\omega\tau\right)$. If the real part of the eigenvalues of $G$ are strictly positive, the spectrum satisfies the following.
    \begin{align*}
    S(\omega)
    &=\frac{1}{2\pi}\left[\left(\frac{1}{2}G-i\omega I\right)^{-1}P_\infty +P_\infty \left(\frac{1}{2}G^{\top}+i\omega I\right)^{-1}\right]\\
    &=\frac{1}{2\pi}\left(\frac{1}{2}G-i\omega I\right)^{-1}NN^{\top}\left(\frac{1}{2}G^{\top}+i\omega I\right)^{-1}    
    \end{align*}
\end{enumerate}
\end{lemma}

\begin{proof} In turn.
\begin{enumerate}
    \item Observe that this matrix equation corresponds to the fixed point equation for the Fokker-Planck equation corresponding to this stochastic differential equation \citep{artemiev1997numerical}.
    \item \cite{vatiwutipong2019alternative} give that $\mathbb{E}[\latV(t)|\latV(0)]=\exp(-Gt/2)\latV(0)$.  Thus
    \[
    \mathbb{E}[\latV(t)\latV(0)^\top] =
    \mathbb{E}[\mathbb{E}[\latV(t)|\latV(0)]\latV(0)] = \exp(-G^\top t/2) P_\infty
    \]
    as desired.
    \item Applying the Fourier inversion formula, we obtain that 
    \[
    S(\omega) = \frac{1}{2\pi} \int_{-\infty}^{\infty} \exp(i\omega \tau)C(\tau)d\tau
    \]
    as long as such an integral converges.  Since $C$ is itself defined differently for $t\geq0$ and $t\leq 0$, it is convenient to split this into parts.  First the positive part.
    \begin{align*}
    \int_{0}^{\infty} e^{i\omega \tau} C(\tau) d\tau 
        &= \left(\int_{0}^{\infty} e^{\tau(i\omega I - G/2)} d\tau\right) P_\infty 
        =  (G/2 - i\omega I)^{-1} P_\infty
    \end{align*}
    Note that this integral does indeed converge, since we have assumed the real part of the eigenvalues of $G$ are strictly positive.  By the same reasoning, the negative part takes the form $P_\infty (G^\top/2 + i\omega I)^{-1}$.  This yields the first expression for $S(\omega)$.  The second expression follows by using part 1 of this Lemma.
\end{enumerate}
\end{proof}

\subsection{Glick's Lemma}

\begin{lemma}[Glick's extension of Scheffe's lemma] \label{lem:glick}

Let $(\Omega, \mathcal{F},\pi)$ a probability measure space.  Let $h_1,h_2,h_3\cdots$ denote a sequence of $\mathcal{F}$-measurable functions of the form $h_\ell:\ \mathbb{R}^d \times \Omega \rightarrow \mathbb{R}$.  Let $h_\infty:\ \mathbb{R}^d \rightarrow\mathbb{R}$ another function.  For Lebesgue-almost-every $x$ and $\pi$-almost-every $\omega$, assume that $\lim_{\ell \rightarrow \infty} h_\ell (x,\omega),h_2(x,\omega) = h_\infty (x)$.  For $\pi$-almost-every $\omega$, assume that  $\lim_{\ell \rightarrow \infty} \int |h_1(x,\omega)|dx = \int |h_\infty(x,\omega)|dx$.  Then, for $\pi$-almost-every $\omega$, we have that
\[
\lim_{\ell \rightarrow \infty} \int |h_\ell (x,\omega)-h_\infty(x)|dx = 0
\]
\end{lemma}
\begin{proof}
    Apply Scheffe's lemma for each $\omega$.  This observation is generally credited to Glick \citep{glick1974consistency}.
\end{proof}

\end{document}